\documentclass[a4paper,fleqn]{cas-dc}

\usepackage[numbers]{natbib}
\usepackage{algorithm}
\usepackage{algorithmic}
\usepackage{amsthm}
\usepackage{caption}
\usepackage{fontawesome5}
\usepackage[edges]{forest}
\usepackage{multicol}
\usepackage{subcaption}
\usepackage{tabularx}
\usepackage{tikz}

\hypersetup{
    hidelinks,
    colorlinks=true,
    linkcolor=blue,
    urlcolor=blue,
    citecolor=blue
}

\definecolor{hidden-draw}{RGB}{20,68,106}
\definecolor{hidden-pink}{RGB}{255,245,247}
\definecolor{lightred}{RGB}{255, 204, 204}
\definecolor{lightgreen}{RGB}{224, 255, 225}
\definecolor{lightyellow}{RGB}{255, 241, 224}
\definecolor{lightpurple}{RGB}{225, 225, 255}
\definecolor{lightgray}{gray}{0.9}
\definecolor{text-red}{RGB}{255, 0, 0}
\definecolor{text-blue}{RGB}{0, 0, 255}
\definecolor{deep-purple}{RGB}{84, 74, 255}
\definecolor{deep-blue}{RGB}{0, 170, 238}
\definecolor{deep-green}{RGB}{63, 183, 4}

\newcommand{\hide}[1]{}
\newcommand{\vpara}[1]{\vspace{0.05in}\noindent\textbf{#1}}

\newcommand{\best}[1]{\textcolor{text-red}{\textbf{#1}}}
\newcommand{\second}[1]{\textcolor{text-blue}{\textbf{#1}}}

\def\model{\textbf{SPDM}}
\renewcommand{\best}[1]{\textbf{#1}}
\renewcommand{\second}[1]{\underline{#1}}

\def\BibTeX{{\rm B\kern-.05em{\sc i\kern-.025em b}\kern-.08em
    T\kern-.1667em\lower.7ex\hbox{E}\kern-.125emX}}

\begin{document}
\let\WriteBookmarks\relax
\def\floatpagepagefraction{1}
\def\textpagefraction{.001}

\shorttitle{SPDM: Geometry-Modulated SSM with Manifold Constraints}

\shortauthors{X. Chen and S.M. Yiu}

\title [mode = title]{SPDM: Geometry-Modulated State Space Modeling with Manifold Constraints for Time Series Forecasting}

\author[1]{Xingsheng Chen}
\affiliation[1]{organization={School of Computing and Data Science, The University of Hong Kong},
    city={Hong Kong SAR},
    country={China}}

\author[1]{Siu-Ming Yiu}

\begin{abstract}
    Multivariate time series forecasting requires capturing the continuously evolving correlation structure among interacting variables. Existing state-space models process time series by scanning tokenized temporal or spatial sequences, discarding the evolutionary geometric structure. We address this limitation by introducing manifold constraints into state-space modeling: treating the cross-variable correlation structure as a continuous trajectory on the symmetric positive definite manifold, whose Riemannian geometric features, tangent space linearity, and Fr\'echet mean centrality act as a principled geometric regularizer that guides and stabilizes the selective scanning dynamics of SSMs. We propose \model, a geometry-aware SSM architecture that realizes this principle through two cooperating mechanisms: a manifold trajectory path that projects dynamically evolving covariance matrices from the SPD manifold to a Euclidean tangent space, and a geometric gating scheme that directly modulates SSM's internal selective parameters based on geometric signals derived from the manifold trajectory. The parameterization preserves the linear-time complexity of the Mamba parallel scan while embedding rich structural constraints, making the architecture preserve prediction accuracy and computational efficiency simultaneously. Extensive experiments on eleven real-world benchmark datasets establish state-of-the-art forecasting performance, and further studies confirm that geometrically constrained state-space dynamics are the dominant architectural factor behind its performance gains. The code is available at~\url{https://github.com/XsChen524/spdm}
\end{abstract}


\maketitle

\section{Introduction}

The ability to model structured dependencies among multiple interacting channels is a fundamental challenge in neural sequence modeling. In complex dynamical systems, whether physical, biological, or manufacturing, multivariate observations continuously record the evolving states of interacting components, yielding time-varying correlation structures that encode how subsystems influence one another and how the system transitions between operational regimes~\cite{PapageorgiouGeorge2025Aswn}. A central challenge therefore lies in equipping neural architectures with the capacity to represent this evolving spatial correlation structure with geometric fidelity, which is absent in both static adjacency-based methods and self-attention mechanisms.

Existing neural sequence architectures largely treat inter-channel relationships as an afterthought. Early statistical and recurrent models treated variables as independent channels, discarding cross-variable dependency information. While the Transformer~\cite{vaswani2017attention} introduced temporal interaction in global range and spurred a series of forecasting frameworks~\cite{liu2023itransformer,zhang2023crossformer}, its $\mathcal{O}(L^2)$ self-attention creates a dual bottleneck: computational infeasibility for long sequences, and a fundamental inability to propagate correlation structure continuously along the temporal dimension. Linearized alternatives~\cite{zeng2023transformers,zhou2022fedformer} reduce overhead but sacrifice temporal precision. Beyond computational considerations, an emerging line of work recognizes that correlation structure possesses intrinsic geometric properties amenable to structured modeling: Koopman-based approaches~\cite{wang2023koopman} learn spectral operators for nonlinear dynamics, TimesNet~\cite{wu2023timesnet} discovers multi-periodicity through 2D temporal restructuring, and HyperTime~\cite{zhang2026multivariate} enforces hyperbolic entailment constraints to embed hierarchical channel relationships in non-Euclidean space. However, these methods either rely on static geometric priors or apply geometric transformations as auxiliary mechanisms, without modeling the correlation structure as a \emph{continuously evolving geometric trajectory} on a principled geometric manifold.

Recently, state-space models (SSMs)~\cite{merrill2024illusion} have emerged as a principled and efficient alternative for sequence modeling. The Mamba architecture~\cite{gu2023mamba} introduces selective scanning mechanisms with linear time complexity, achieving superior scalability over Transformer-based methods, and has been extended to time series forecasting through bidirectional scanning~\cite{liang2024bi,wang2025mamba} and inverted embedding~\cite{liu2023itransformer}. However, existing Mamba-based models mainly perform modeling and prediction through causal scanning on varied dimensions, while even graph-based methods that explicitly model pairwise variable dependencies~\cite{wang2024contrastive} treat the correlation structure as a static graph, both discarding the geometric continuity essential for capturing its temporal evolution. DyGraphformer~\cite{han2025dygraphformer} employs dynamic graph convolution to infer time-varying spatial dependencies, yet still relies on discrete adjacency structures within a Transformer backbone. RCL~\cite{yan2025repetitive} highlights that standard Mamba blocks suffer from insufficient focus on critical time steps, motivating architectures that strengthen selectivity through principled geometric guidance rather than post-hoc pretraining.

To address these limitations, we introduce the principle of manifold-constrained state-space dynamics. The central insight is that the evolving cross-variable correlation structure resides on the manifold of symmetric positive definite (SPD) matrices $\mathcal{P}_N$, whose geodesic smoothness, tangent space linearity, and Fr\'echet mean centrality provide principled geometric regularization for SSM-based forecasting. We realize this principle through \textbf{\model}, which models the cross-variable correlation structure as a continuous trajectory on $\mathcal{P}_N$, scans it with a dedicated Geometry Mamba, and directly injects resulting geometric signals into the selective parameters of a bidirectional Mamba Inter-variate Correlations (VC) Encoder. In this way it effectively couples the SSM's information gating to the evolving manifold geometry. In contrast to existing approaches that impose static geometric constraints~\cite{zhang2026multivariate,li2026dfim} or model variable relationships as discrete graph structures, \model\ explicitly intervenes at the SSM parameterization level, making geometry a formative component of the sequence modeling process.

Our primary contributions are:
\begin{itemize}
    \item We introduce the principle of manifold-constrained state-space dynamics, demonstrating that treating the evolving cross-variable correlation structure as a continuous Riemannian trajectory on the SPD manifold provides effective geometric regularization for SSM-based multivariate forecasting.
    \item We propose \model, a geometry-aware SSM architecture that realizes this principle by scanning SPD manifold trajectories with a Geometry Mamba in the tangent space and directly modulating Mamba's selective parameters through geometric gating, maintaining linear time complexity while achieving intrinsic geometric awareness without external conditioning.
    \item We provide comprehensive experimental evidence by conducting systematic studies on eleven real-world benchmark datasets, confirming that manifold-constrained dynamics constitute the dominant factor behind \model's state-of-the-art forecasting performance instead of isolated architectural components.
\end{itemize}

Through modeling state-space for the given sequences in non-Euclidean space and modulating the intrinsic parameters of the scanning process, \model\ achieves efficient, geometric cross-variate pattern recognition, as well as maintaining considerable scalability given sequences of increasing lengths. This work establishes a foundation for manifold-constrained neural dynamics in state-space architectures, opening avenues for geometric deep learning on structured non-Euclidean representations.

\section{Related Work}

\subsection{Mamba and State Space Models for Time Series}
SSMs have recently emerged as an efficient paradigm for long-sequence modeling. The Mamba architecture~\cite{gu2023mamba} introduces a selective scanning mechanism that enables input-dependent state transitions with linear-time complexity, offering significant efficiency advantages over attention-based models. For time series forecasting, S-Mamba~\cite{wang2025mamba} employs bidirectional Mamba branches to capture forward and reverse temporal dependencies, while BiMamba4TS~\cite{liang2024bi} extends this with specialized architectures for multivariate settings. RCL~\cite{yan2025repetitive} enhances Mamba's selectivity for time series prediction through token-level contrastive pretraining, highlighting that standard Mamba blocks suffer from insufficient focus on critical time steps, a limitation that motivates our architectural intervention at the SSM parameterization level. Beyond time series, Mamba variants have been adopted across diverse domains, including FA-Mamba~\cite{yang2026fa} for multimodal remote sensing classification and Geo-Mamba~\cite{Cao2026GeoMamba} for geometry-informed functional brain organization learning, illustrating the broad applicability of the Mamba architectural paradigm. However, existing Mamba-based methods primarily focus on one-dimensional temporal scanning and treat multivariate interactions as flat sequences. They lack explicit mechanisms to capture the \emph{geometric structure} of evolving inter-variable correlations, which is the core focus of \model.

\subsection{Riemannian Geometry for Multivariate Modeling}
The application of Riemannian geometry to multivariate data analysis has a rich history. The SPD manifold $\mathcal{P}_N$ arises naturally when modeling covariance or correlation matrices, with applications spanning brain-computer interfaces, action recognition, and medical imaging. Foundational work on Riemannian geometry of matrix geometric means~\cite{bhatia2006riemannian} established key properties of geodesics and barycenters on $\mathcal{P}_N$, while the log-Euclidean framework~\cite{arsigny2007geometric} defines a family of Riemannian metrics on $\mathcal{P}_N$ by embedding the manifold in a Euclidean space through the matrix logarithm, offering a computationally efficient alternative to the affine-invariant metric while preserving key geometric properties such as invariance under inversion and similarity transformations. Classical methods such as geodesic filtering and Fr\'echet mean computation on $\mathcal{P}_N$ provide principled tools for statistical analysis of covariance trajectories.

In the broader context of manifold-based time series analysis, Slavakis et al.~\cite{slavakis2017clustering} pioneered Riemannian multi-manifold modeling for brain-network time series, viewing temporal features as points on a union of Riemannian submanifolds; this line of work was extended by Ye et al.~\cite{ye2021fast} to fast sequential clustering on dynamic multilayer networks. RM-BLS~\cite{feng2019robust} utilizes manifold embedding for robust chaotic system prediction through random perturbation approximation, further demonstrating the value of manifold representations for sequential data. More recently, Riemannian deep learning has extended neural network operations to manifold-valued data, enabling end-to-end learning on SPD matrices. Nava-Yazdani~\cite{navayazdani2025ridge} proposed ridge regression directly on Riemannian manifolds for time-series prediction, providing explicit gradient formulations for manifold-valued regression via B\'{e}zier curves. However, these geometric methods have not been integrated with efficient state-space scanning for time series forecasting, a gap that \model\ directly addresses by projecting SPD trajectories to the tangent space and leveraging Mamba's parallel scan.

\subsection{Geometric Deep Learning on Manifolds}
Geometric deep learning provides a unified framework for generalizing neural networks to non-Euclidean domains. For the SPD manifold specifically, SPDNet~\cite{huang2017spdnet} and its variants have introduced layers that respect the Riemannian geometry, such as BiMap, ReEig, and LogEig operations. Tangent space methods have been employed to map manifold-valued features to Euclidean spaces where standard architectures can operate. Recently, ManifoldFormer~\cite{fu2026manifoldformer} extended this paradigm through geodesic-aware attention and manifold-constrained neural ODEs for neural dynamics modeling on Riemannian manifolds. Complementary theoretical studies have employed Riemannian curvature tensors to characterize trained deep networks~\cite{kaul2019riemannian} and geometry flow to preserve data manifold structure in deep metric learning~\cite{li2023geometry}. Recent work on Riemannian SSMs has explored manifold-constrained recurrent dynamics, but efficient parallel scanning on learned manifold trajectories remains underexplored. \model\ bridges this gap by injecting the output of Geometry Mamba scanning on trajectory into the mature Mamba VC encoder, achieving both geometric fidelity and computational efficiency.

\subsection{Organization}
The remainder of this paper is organized as follows. Section~\ref{sec:methodology} presents the \model\ framework, including the manifold trajectory construction, tangent space projection, geometric gating mechanism, and prediction pipeline. Section~\ref{sec:experiments} reports comprehensive experiments on eleven benchmark datasets, covering forecasting effectiveness, ablation studies, efficiency analysis, etc. Section~\ref{sec:conclusion} concludes the work.

\begin{figure*}[!t]
    \centering
    \includegraphics[width=\linewidth]{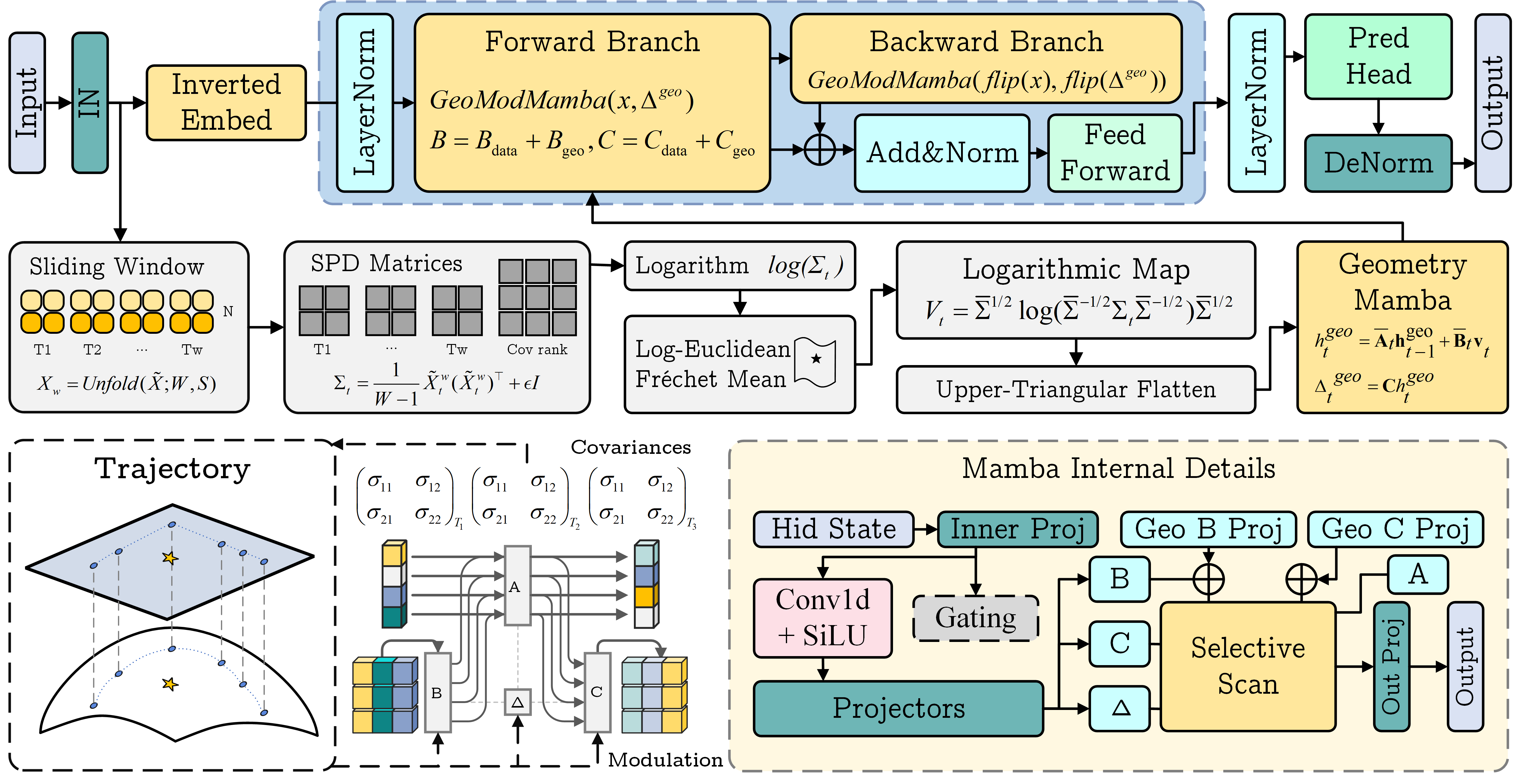}
    \caption{Framework of~\model}
    \label{fig:framework}
\end{figure*}

\section{Methodology}
\label{sec:methodology}

\subsection{Problem Formulation}

Let $\mathbf{X} = [\mathbf{x}_1, \mathbf{x}_2, \ldots, \mathbf{x}_L] \in \mathbb{R}^{L \times N}$ denote a multivariate time series with $L$ historical time steps and $N$ variables. Each vector $\mathbf{x}_t = [x_t^{1}, x_t^{2}, \ldots, x_t^{N}]$ represents the observations of $N$ correlated signals at time step $t$. Given a historical observation window of length $L$, the forecasting objective is to predict the next $P$ time steps:
\begin{equation}
    \hat{\mathbf{Y}} = [\hat{\mathbf{x}}_{L+1}, \hat{\mathbf{x}}_{L+2}, \ldots, \hat{\mathbf{x}}_{L+P}] = f_\theta(\mathbf{X}_{1:L})
\end{equation}
where $f_\theta(\cdot)$ represents a learnable forecasting model parameterized by $\theta$, and $\hat{\mathbf{Y}} \in \mathbb{R}^{P \times N}$ is the predicted future sequence.

In multivariate forecasting, a key challenge is capturing evolving temporal and spatial dynamics in given time windows. At each time step $t$, the cross-variable covariance matrix $\Sigma_t$ encodes the instantaneous correlation patterns; as time advances, the sequence $\{\Sigma_1, \ldots, \Sigma_T\}$ traces a trajectory that reflects how inter-variable dependencies evolve. Traditional approaches either model this as a static graph or rely on attention mechanisms with quadratic complexity, failing to capture the continuous geometric nature of this evolution.

\subsection{Overall Architecture}

\model\ models the evolving spatiotemporal correlation structure as a \textbf{continuous Riemannian trajectory} on the SPD manifold $\mathcal{P}_N$. The core insight is that each covariance matrix $\Sigma_t$ of a time window naturally resides on $\mathcal{P}_N$, and projecting this trajectory to a Euclidean tangent space enables seamless integration with the efficient Mamba parallel scan. As illustrated in Figure~\ref{fig:framework}, the model comprises two complementary paths:

\textbf{Path A: SPD Manifold Trajectory Path} constructs a sliding-window dynamic covariance sequence on $\mathcal{P}_N$, computes the log-Euclidean Fr\'echet mean as a reference point, projects each covariance matrix to the tangent space via the logarithmic map, flattens the resulting symmetric matrices into vectors, and processes the tangent vector sequence through a dedicated Geometry Mamba to produce the modulation parameter $\Delta^{\text{geo}}$.

\textbf{Path B: Modulated Mamba Encoder} processes sequences through inverted embedding and a stack of bidirectional Mamba VC encoder layers. Crucially, the geometric signals from Path A are injected directly into the selective state-space parameters ($B$, $C$ by default) of Mamba VC through dedicated learned projections $\mathbf{W}_B^{\text{geo}}$ and $\mathbf{W}_C^{\text{geo}}$, making it feasible to modulate historical sequence processing and prediction with geometric features.

This modulation is not merely a data transformation but a principled mechanism for regularizing the latent space of the SSM: by characterizing and constraining the state-space dynamics on the SPD manifold, we effectively incorporate the geometric consistency of data into forecasting while maintaining remarkable scalability. The full pipeline is summarized in Algorithm~\ref{alg:spdm_pipe}. Before entering the two paths, the input $\mathbf{X} \in \mathbb{R}^{B \times L \times N}$ is normalized along the temporal dimension: $\tilde{\mathbf{X}} = (\mathbf{X} - \boldsymbol{\mu}) / \boldsymbol{\sigma}$, where $\boldsymbol{\mu}, \boldsymbol{\sigma} \in \mathbb{R}^N$ are computed per-variable with population variance. Then the sequences are passed to separate paths, one of which is to model the geometric information and generate the modulation parameters, while the other one is to scan and modeling the spatial pattern.

\subsection{Path A: SPD Manifold Trajectory}
\label{sec:path_a}

Path A constructs and processes the Riemannian trajectory of inter-variable correlations on the SPD manifold. The pipeline proceeds through five stages, each designed to preserve geometric fidelity while enabling efficient computation.

\subsubsection{Dynamic Covariance Construction}

\begin{figure*}[!t]
    \centering
    \includegraphics[width=\linewidth]{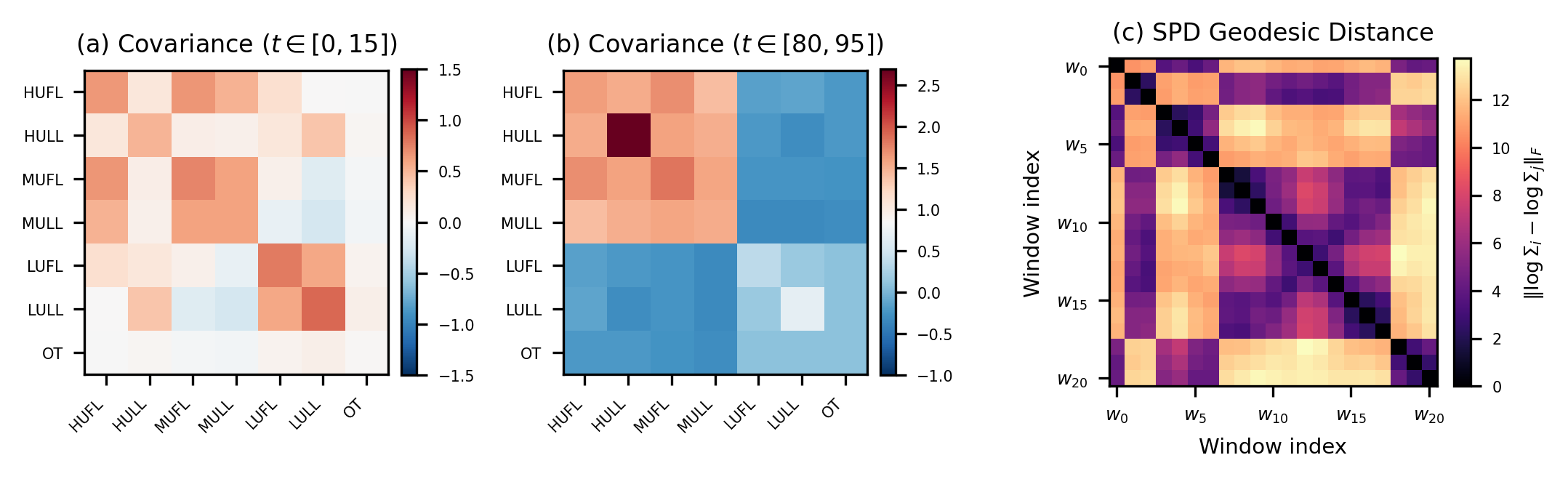}
    \caption{SPD covariance geometry along the input sequence on ETTm2 with 7 variables. (a) Sample covariance matrix $\Sigma_0$ at the first window ($t \in [0, 15]$), exhibiting moderate variances with diagonal range 0.01-0.87 and predominantly positive cross-variable correlations among the high and mid-frequency load series (HUFL-MULL). (b) Covariance matrix $\Sigma_{20}$ at the last window ($t \in [80, 95]$). The variances of HUFL-MULL increase by up to $5\times$ compared to (a), also their mutual correlations strengthen dramatically. The negative cross-group correlations emerge with the low-frequency variables, indicating a substantial regime shift in the underlying multivariate dynamics. (c) Pairwise Log-Euclidean geodesic distance matrix $d(\Sigma_i, \Sigma_j) = \|\log \Sigma_i - \log \Sigma_j\|_F$ across all windows. The structure reveals distinct covariance regimes separated by abrupt transitions, rather than a smooth drift. Notably, the late windows ($w_{18}$-$w_{20}$) return geometrically close to the early windows ($w_0$, $w_3$-$w_6$; $d \approx 3$-$5$), suggesting near-periodic dynamics in the covariance structure. These non-stationary patterns in the evolving covariance geometry motivate the SPD tangent projection pathway.}
    \label{fig:exp_1}
\end{figure*}

Given the normalized input $\tilde{\mathbf{X}}$, we first unfold the sequence along time dimension into overlapping sliding windows to obtain local snapshots of the correlation structure:
\begin{equation}
    \mathbf{X}_w = \text{Unfold}(\tilde{\mathbf{X}}; W, S) \in \mathbb{R}^{B \times T_w \times N \times W}
\end{equation}
where $W$ is the window size, $S$ is the stride, and $T_w = \lfloor (L - W) / S \rfloor + 1$ is the number of windows. For datasets with a large number of variables (e.g., ECL with $N = 321$), an optional low-rank projection $\mathbf{W}_{\text{proj}} \in \mathbb{R}^{R \times N}$ with $R \ll N$ reduces the effective variable dimension from $N$ to $R$. For each window position $t \in \{1, \ldots, T_w\}$, the instantaneous spatial correlation is captured by the sample covariance matrix as demonstrated in figure~\ref{fig:exp_1}:
\begin{equation}
    \Sigma_t = \frac{1}{W - 1} \tilde{\mathbf{X}}_w^t (\tilde{\mathbf{X}}_w^t)^\top + \epsilon \mathbf{I} \quad \in \mathcal{P}_N
\end{equation}
where $\tilde{\mathbf{X}}_w^t \in \mathbb{R}^{N \times W}$ is the mean-centered window content, $\epsilon$ is a small regularization constant (e.g., $10^{-4}$) ensuring positive definiteness, and $\mathbf{I}$ is the identity matrix. The sequence $\{\Sigma_1, \ldots, \Sigma_{T_w}\}$ traces a discrete trajectory on manifold $\mathcal{P}_N$.

\subsubsection{Tangent Space Projection}

To project the trajectory to a flat Euclidean space amenable to Mamba block processing, we first establish a geometrically principled reference point on $\mathcal{P}_N$. The Fr\'echet mean, Riemannian center of mass, under the log-Euclidean metric admits a closed-form solution:
\begin{equation}
    \bar{\Sigma} = \exp\left(\frac{1}{T_w} \sum_{t=1}^{T_w} \log(\Sigma_t)\right) \quad \in \mathcal{P}_N
\end{equation}
where $\log(\cdot)$ and $\exp(\cdot)$ are the matrix logarithm and matrix exponential computed via eigenvalue decomposition. Unlike the arithmetic mean used in prior work, the Fr\'echet mean respects the curved geometry of $\mathcal{P}_N$, avoiding projection distortions when covariance matrices differ substantially. With the reference point $\bar{\Sigma}$ established, each covariance matrix $\Sigma_t$ is projected to the tangent space $T_{\bar{\Sigma}} \mathcal{P}_N$ via the logarithmic map:
\begin{equation}
    \mathbf{V}_t = \log_{\bar{\Sigma}}(\Sigma_t) = \bar{\Sigma}^{1/2} \; \log\left(\bar{\Sigma}^{-1/2} \, \Sigma_t \, \bar{\Sigma}^{-1/2}\right) \; \bar{\Sigma}^{1/2}
\end{equation}
where $\bar{\Sigma}^{1/2}$ and $\bar{\Sigma}^{-1/2}$ are the matrix square-root and its inverse, reused from the Fr\'echet mean eigendecomposition. All matrix operations (log, exp, sqrt, inv-sqrt) are computed via eigendecomposition, with eigenvalues clamped to $[\epsilon, \infty)$ for numerical stability. The resulting tangent vectors $\mathbf{V}_t \in T_{\bar{\Sigma}} \mathcal{P}_N$ are symmetric matrices lying in a Euclidean vector space. To manage the $\mathcal{O}(N^3)$ cost of eigendecompositions for datasets with numerous variables, previously mentioned low-rank projection reduces the effective dimension to $R \ll N$, lowering the per-eigendecomposition cost to $\mathcal{O}(R^3)$.

Since each tangent vector $\mathbf{V}_t \in \mathbb{R}^{N \times N}$ is symmetric, we extract only its upper-triangular elements to eliminate redundancy by $\mathbf{v}_t = \text{triu}(\mathbf{V}_t) \in \mathbb{R}^{N(N+1)/2}$. The resulting vector $\mathbf{v}_t$ encodes the evolution of spatial correlation structure as a trajectory of flat vectors, where each dimension corresponds to a specific pair of variables' correlation dynamics.

\subsubsection{Geometry Mamba}

The tangent vector sequence is processed by a Geometry Mamba module to capture the temporal dynamics of the geometric trajectory:
\begin{equation}
    \mathbf{h}_{\text{geo}} = \text{LN}(\text{Mamba}(\mathbf{W}_{\text{in}} \mathbf{v}) + \mathbf{W}_{\text{in}} \mathbf{v}) \in \mathbb{R}^{B \times T_w \times d_{\text{geo}}}
\end{equation}
where $\mathbf{W}_{\text{in}} \in \mathbb{R}^{N(N+1)/2 \times d_{\text{geo}}}$ projects the flattened upper-triangular vector to the geometry model dimension. Given the short tangent sequence length (equal to $T_w$), the Geometry Mamba operates with compact dimensions to maintain its inferential ability and avoid involving noise. Its output $\mathbf{h}_{\text{geo}}$ encodes the evolving geometric state of inter-variable correlations, helping with spatial state-space modeling.

\subsubsection{Sparse Window Alignment}

The Geometry Mamba output $\mathbf{h}_{\text{geo}} \in \mathbb{R}^{B \times T_w \times d_{\text{geo}}}$ encodes the SPD trajectory at window granularity, while the Mamba VC encoder in Path~B operates at the variable token level ($N$ positions). To bridge this granularity gap, a \textbf{Sparse Window Alignment} strategy is adopted:
\begin{equation}
    \boldsymbol{\Delta}_{\text{geo}}^{[i]} = \begin{cases}
        \mathbf{W}_{\text{proj}} \mathbf{h}_{\text{geo}}^{[w]}, & \text{if } i = w \cdot S \text{ for some } w \in [0, T_w) \\
        \mathbf{0},                                             & \text{otherwise}
    \end{cases}
\end{equation}
where $\mathbf{W}_{\text{proj}} \in \mathbb{R}^{d_{\text{geo}} \times d_{\text{inner}}}$ projects geometry features to the Mamba inner dimension. The sparse design is well-motivated: each covariance matrix $\Sigma_t$ characterizes correlations over a contiguous window, and spreading its tangent vector across positions via interpolation would mix distinct covariance regimes, diluting the geometric signal. Sparse alignment instead ensures each variable token receives the geometric context most causally relevant to its temporal neighborhood, preserving the native resolution of the SPD trajectory. Ablation experiments (Section~\ref{sec:ablation}) confirm that this approach consistently outperforms linear interpolation alternatives.

\subsection{Path B: Mamba VC Encoder with Modulation}
\label{sec:path_b}

Path B processes the normalized time series through Mamba VC encoder layers whose selective state-space dynamics are adaptively modulated by geometric signals from Path A. Following the inverted embedding operation, sequences of variates are treated as tokens that encode full temporal history into feature vectors of model dimension. Modulated Mamba scans bidirectionally along the variable channel to capture cross-variable dependencies, propagating to feed-forward network (FFN) layer for refinement.

\subsubsection{Inverted Embedding}

Following the inverted embedding paradigm~\cite{liu2023itransformer}, each variable is treated as a token. The normalized input $\tilde{\mathbf{X}} \in \mathbb{R}^{B \times L \times N}$ is transposed so that the variable axis becomes the primary sequence dimension, enabling the Mamba to scan across variables rather than time steps. The transposed sequence is then projected to the model dimension:
\begin{equation}
    \mathbf{H}_{\text{enc}} = \text{Linear}_{L \to d}(\tilde{\mathbf{X}}^\top) \in \mathbb{R}^{B \times N \times d}
\end{equation}
Auxiliary time-feature tokens (e.g., month, day, weekday, hour) provided in $X_{\text{mark}}$ for specific datasets are also concatenated as additional variable tokens, bringing the total token count to $N = N_{\text{vars}} + M$.

\subsubsection{Geometry-Modulated Mamba}

The core innovation of \model\ lies in the \textbf{Geometry-Modulated Mamba} module, which directly injects geometric signals from Path A into the selective state-space parameters of the Mamba VC. Rather than treating the geometric trajectory as an external conditioning signal applied to the Mamba output, Geometry-Modulated Mamba intercepts the internal SSM parameterization to make the selective scan itself geometry-aware. By modulating $\mathbf{B}$ and $\mathbf{C}$, extracted manifold constraint information directly influences SSM's input-to-state and state-to-output pathways, transforming the selective scan from a purely causal inference mechanism into a topology-aware neural primitive.

A standard Mamba selective scan operates as follows: the input $\mathbf{x} \in \mathbb{R}^{L \times d}$ is projected through $\mathbf{W}_{\text{in}}$ and split into $\mathbf{x}, \mathbf{z} \in \mathbb{R}^{d_{\text{inner}}}$; after depthwise convolution and activation, $\mathbf{x}$ is projected by $\mathbf{W}_{\text{x}}$ to produce the time-step size $\boldsymbol{\Delta} \in \mathbb{R}^{d_{\text{inner}}}$ (via $\mathbf{W}_{\Delta}$ with rank $r$) and the input/output projection parameters $\mathbf{B}, \mathbf{C} \in \mathbb{R}^{S}$; the selective scan then computes:
\begin{equation}
    \mathbf{h}_t = \exp(\boldsymbol{\Delta}_t \mathbf{A}) \mathbf{h}_{t-1} + \boldsymbol{\Delta}_t \mathbf{B} \mathbf{x}_t, \quad \mathbf{y}_t = \mathbf{C} \mathbf{h}_t + \mathbf{D} \mathbf{x}_t
\end{equation}
where $\mathbf{A} \in \mathbb{R}^{d_{\text{inner}} \times S}$ is a diagonal state transition matrix (initialized as $\text{diag}(1, 2, \ldots, S)$), $\mathbf{B}, \mathbf{C} \in \mathbb{R}^{S}$ are the input and output projection vectors, $\mathbf{D} \in \mathbb{R}^{d_{\text{inner}}}$ is a residual connection, and $\exp(\boldsymbol{\Delta}_t \mathbf{A})$ is element-wise exponentiation.

Geometry-Modulated Mamba intercepts SSM parameterization \emph{after} the $\mathbf{x}$-projection produces $\mathbf{B}^{\text{base}}$ and $\mathbf{C}^{\text{base}}$. The geometric signal $\boldsymbol{\Delta}_{\text{geo}} \in \mathbb{R}^{B \times N \times d_{\text{inner}}}$ is projected through learned transformations $\mathbf{W}_B^{\text{geo}}, \mathbf{W}_C^{\text{geo}} \in \mathbb{R}^{d_{\text{inner}} \times S}$ and additively injected into the SSM input/output projection parameters, where each projection comprises a two-layer MLP with GELU activation:
\begin{equation}
    \mathbf{B} = \mathbf{B}^{\text{base}} + \mathbf{W}_B^{\text{geo}}(\boldsymbol{\Delta}_{\text{geo}}), \quad
    \mathbf{C} = \mathbf{C}^{\text{base}} + \mathbf{W}_C^{\text{geo}}(\boldsymbol{\Delta}_{\text{geo}})
\end{equation}
This design has a clear physical interpretation: by modulating $\mathbf{B}$ and $\mathbf{C}$, the geometric trajectory directly controls which information enters and exits SSM's hidden state $\mathbf{h}_t$ and to what extent. When the covariance structure undergoes a shift, the geometric signal amplifies the input projection for variables whose correlation patterns are changing, while suppressing contributions from stable regions. The additive injection is applied identically in both the forward and reverse Mamba directions, ensuring bidirectional geometric awareness.

Consequently, the geometric gating constitutes an architectural intervention at the SSM parameterization level: it couples the Mamba's selection mechanism to the evolving covariance geometry of the input distribution, enabling the latent state $\mathbf{h}_t$ to adapt its information retention and retrieval dynamics in response to structural changes in the underlying recognized pattern. Alternative injection designs (bounded scaling, $\Delta$ modulation) and alignment strategies are evaluated in Section~\ref{sec:ablation}.

\begin{algorithm}[tb]
	\caption{Forecasting Pipeline of \model}
	\label{alg:spdm_pipe}
	\begin{algorithmic}
		\REQUIRE \textbf{Input} $X \in \mathbb{R}^{B \times L \times N}$, $X_{mark} \in \mathbb{R}^{B \times L \times M}$
		\STATE $\tilde{X} \leftarrow \text{InstanceNorm}(X)$
		\STATE $\triangleright$ \textbf{Path A: SPD Manifold Trajectory}
		\STATE $X_w \leftarrow \text{Unfold}(\tilde{X}; W, S)$ \hfill $\triangleright [B, T_w, N, W]$
		\IF{Low-rank ($R > 0$)}
			\STATE $X_w \leftarrow \text{Linear}_{N \to R}(X_w)$ \hfill $\triangleright [B, T_w, R, W]$
		\ENDIF
		\STATE $D \leftarrow R$ if low-rank else $N$
		\STATE $\Sigma_t \leftarrow \frac{1}{W-1} \tilde{X}_w^t (\tilde{X}_w^t)^\top + \epsilon \mathbf{I}$ \hfill $\triangleright [B, T_w, D, D]$
		\STATE $\bar{\Sigma} \leftarrow \text{MatrixExp}\big(\text{mean}(\log(\Sigma_t))\big)$ \hfill $\triangleright$ Log-Euclidean Fr\'echet mean; $\bar{\Sigma}$, $\mathbf{V}$, $\boldsymbol{\lambda}$ returned jointly
		\STATE $\bar{\Sigma}^{1/2} \leftarrow \mathbf{V} \text{diag}(\sqrt{\boldsymbol{\lambda}}) \mathbf{V}^\top$, $\bar{\Sigma}^{-1/2} \leftarrow \mathbf{V} \text{diag}(1/\sqrt{\boldsymbol{\lambda}}) \mathbf{V}^\top$ \hfill $\triangleright$ Reuses eigendecomposition from Fr\'echet mean
		\STATE $\mathbf{V}_t \leftarrow \bar{\Sigma}^{1/2} \cdot \log(\bar{\Sigma}^{-1/2} \Sigma_t \bar{\Sigma}^{-1/2}) \cdot \bar{\Sigma}^{1/2}$ \hfill $\triangleright$ Logarithmic map
		\STATE $\mathbf{v}_t \leftarrow \text{triu}(\mathbf{V}_t)$ \hfill $\triangleright [B, T_w, D(D+1)/2]$
		\STATE $\mathbf{h}_{\text{geo}} \leftarrow \text{GeometryMamba}(\mathbf{v})$ \hfill $\triangleright [B, T_w, d_{\text{geo}}]$
		\STATE $\triangleright$ \textbf{Sparse Window Alignment}
		\STATE $\boldsymbol{\Delta}_{\text{geo}} \leftarrow \text{SparseAlign}(\mathbf{W}_{\text{proj}} \cdot \mathbf{h}_{\text{geo}})$ \hfill $\triangleright [B, N, d_{\text{inner}}]$
		\STATE $\triangleright$ \textbf{Path B: Mamba VC Encoder}
		\STATE $\mathbf{H}_{\text{enc}} \leftarrow \text{DataEmbed\_inv}(\tilde{X}, X_{mark})$ \hfill $\triangleright [B, N, d]$
		\FOR{$l = 1$ to $E$}
			\STATE $\mathbf{O} \leftarrow \text{GMamba}_{\text{fwd}}(\mathbf{H}_{\text{enc}}, \boldsymbol{\Delta}_{\text{geo}})$ \hfill $\triangleright$ $B$ + $C$ geometric injection
			\STATE $\mathbf{O} \mathrel{+}= \text{Flip}(\text{GMamba}_{\text{rev}}(\text{Flip}(\mathbf{H}_{\text{enc}}), \text{Flip}(\boldsymbol{\Delta}_{\text{geo}})))$
			\STATE $\mathbf{H}_{\text{enc}} \leftarrow \text{LayerNorm}(\mathbf{H}_{\text{enc}} + \mathbf{O})$
			\STATE $\mathbf{H}_{\text{enc}} \leftarrow \text{LayerNorm}(\mathbf{H}_{\text{enc}} + \text{FFN}(\mathbf{H}_{\text{enc}}))$
		\ENDFOR
		\STATE $\mathbf{H}_{\text{enc}} \leftarrow \text{LayerNorm}(\mathbf{H}_{\text{enc}})$
		\STATE $\triangleright$ \textbf{Prediction}
		\STATE $\hat{Y} \leftarrow \text{Linear}_{d \to P}(\mathbf{H}_{\text{enc}}^{[:,:N,:]})$ \hfill $\triangleright [B, N, P]$
		\STATE $\hat{Y} \leftarrow \text{Denorm}(\hat{Y})$
		\STATE \textbf{Output} $\hat{Y} \in \mathbb{R}^{B \times P \times N}$
	\end{algorithmic}
\end{algorithm}


\subsubsection{Bidirectional Mamba VC Encoding}

Multiple encoder layers are stacked, with each of them applying Geometry-Modulated Mamba in both forward and reverse directions along the variable channel for bidirectional cross-variable coverage. The Mamba VC employs a compact depthwise convolution kernel since the spatial dimension lacks the smooth continuity that benefits from larger kernels. The bidirectional scanning and modeling can be expressed as:
\begin{equation}
    \mathbf{O} = \text{GMamba}_{\text{fwd}}(\mathbf{H}) + \text{Flip}\big(\text{GMamba}_{\text{rev}}(\text{Flip}(\mathbf{H}))\big)
\end{equation}
where $\text{Flip}(\cdot)$ reverses the order of variable tokens, enabling the SSM to aggregate cross-variable information from both directions. The bidirectional output is combined with a residual connection, followed by LayerNorm and an FFN with GELU activation:
\begin{equation}
    \mathbf{H}' = \text{LN}(\mathbf{H} + \mathbf{O}), \quad \mathbf{H}'' = \text{LN}(\mathbf{H}' + \text{FFN}(\mathbf{H}'))
\end{equation}
where $\text{FFN}$ applies dense non-linear connections to refine the cross-variable representations of each token. The residual connections preserve gradient flow and enable identity-preserving information propagation through deep layers, while LayerNorm enhances convergence and training stability by standardizing activations to a Gaussian distribution.

\subsection{Prediction}

After a stack of encoder layers, a concluding LayerNorm standardizes the latent representations $\mathbf{H}_{\text{enc}} \in \mathbb{R}^{B \times N' \times d}$ and ensures stable activation statistics entering the prediction head. Then a linear projector maps each variable token's representation to the full prediction horizon:
\begin{equation}
    \hat{\mathbf{Y}} = \text{Linear}_{d \to P}(\mathbf{H}_{\text{enc}}) \in \mathbb{R}^{B \times N \times P}
\end{equation}
A permute operation restores the conventional time-first layout $[B, P, N]$ from inverted embedding diagram. Finally, a de-normalization recovers the original data scale with $\hat{\mathbf{Y}}' = \hat{\mathbf{Y}} \cdot \boldsymbol{\sigma} + \boldsymbol{\mu} \in \mathbb{R}^{B \times P \times N}$. The operations in the prediction phase ensure \model\ outputs predictions with considerable efficiency and accuracy, while avoiding computational bottlenecks or factors that interfere with the training process.

\subsection{Model Complexity Analysis}

The manifold-constrained SSM framework provides theoretical guarantees for stable state propagation: since the Fr\'echet mean of a set of SPD matrices remains on $\mathcal{P}_N$, and the tangent space is isomorphic to a Euclidean vector space, the discretized SSM state transitions enjoy the same numerical stability properties as standard Mamba, with geometry-induced perturbations bounded by the spectral radius of the projected tangent vectors.

Path A (SPD Trajectory) involves sliding-window covariance construction costing $\mathcal{O}(T_w \cdot D^2 \cdot W)$ where $D = N$ without low-rank or $D = R$ with low-rank projection, and $2T_w+1$ eigendecompositions on $D \times D$ matrices costing $\mathcal{O}(T_w \cdot D^3)$. Overall Path A complexity is $\mathcal{O}(T_w \cdot D^2 \cdot W + T_w \cdot D^3)$, reducible to $\mathcal{O}(T_w \cdot R^3)$ via low-rank projection. Path B (Mamba VC Encoder) costs $\mathcal{O}(E \cdot N \cdot d \cdot (d_{\text{inner}} + d_{\text{ff}}))$, which is linear in the variable count $N$ and independent of the original temporal length $L$, with the geometric injection projections adding $\mathcal{O}(N \cdot d_{\text{inner}} \cdot S)$ per layer. Overall, \model\ maintains linear complexity with respect to all key dimensions, avoiding $\mathcal{O}(L^2)$ attention costs. The selective scan leverages a work-efficient parallel associative scan with $\mathcal{O}(\log N)$ parallel depth and $\mathcal{O}(N \cdot d_{\text{inner}} \cdot S)$ total work.

\section{Experiments}
\label{sec:experiments}

We conduct experiments to answer the following research questions:
\textbf{RQ1:} How does \model\ compare against state-of-the-art baselines in overall forecasting accuracy?
\textbf{RQ2:} What is the contribution of each architectural component to \model's performance?
\textbf{RQ3:} How does \model\ balance forecasting accuracy with computational efficiency?
\textbf{RQ4:} How robust is \model\ under varying levels of input noise?
\textbf{RQ5:} How does varying the lookback window length affect \model's forecasting accuracy?
\textbf{RQ6:} How sensitive is \model's performance to variations in key hyperparameters?

\subsection{Experimental Setup}

All experiments were conducted on an Ubuntu 24.04 server with NVIDIA RTX 4090/5090 GPUs, running Python 3.12, CUDA 12.8, PyTorch 2.8.0, and Mamba 2.3.1. We evaluate \model\ using eleven publicly available benchmark datasets as summarized in Table~\ref{tab:dataset_overview}. These datasets span diverse application domains, variable counts (from 7 to 883), and temporal granularities (from 5 minutes to 1 week), providing a comprehensive testbed for evaluating both forecasting accuracy and scalability. MSE and MAE are used as evaluation metrics.

We compare \model\ against eight leading models spanning diverse architectural paradigms: \textbf{S-Mamba}~\cite{wang2025mamba}, a bidirectional Mamba with inverted embedding. \textbf{iTransformer}~\cite{liu2023itransformer} treats each variate as a token through inverted embedding with self-attention; \textbf{PatchTST}~\cite{huang2024long}, which segments sequences into patches for channel-independent modeling; and \textbf{Crossformer}~\cite{zhang2023crossformer}, which employs a two-stage attention mechanism across time and variable dimensions. Additional baselines include \textbf{TiDE}~\cite{das2023long}, a simple MLP-based encoder-decoder with dense residual connections; \textbf{TimesNet}~\cite{wu2023timesnet}, which transforms 1D time series into 2D tensors for multi-periodicity modeling; \textbf{DLinear}~\cite{zeng2023transformers}, a decomposition-based linear model that serves as a competitive simplicity baseline; and \textbf{Autoformer}~\cite{wu2021autoformer}, a Transformer with auto-correlation mechanism for series decomposition. Beyond the main effectiveness comparison, we additionally evaluate \textbf{BiMamba4TS}~\cite{liang2024bi} (bidirectional Mamba with channel-independent patching) and \textbf{interPDN}~\cite{kong2025interpdn} (patch-based decomposition with RevIN and contrastive learning) in the efficiency comparison (Section~\ref{sec:efficiency}) and the lookback length study (Section~4.6).

\begin{table}[htbp]
    \vspace{-0.1in}
    \centering
    \caption{Overview of 11 publicly available datasets.}\vspace{0pt}
    \label{tab:dataset_overview}
    \resizebox{0.98\linewidth}{!}{
        \renewcommand{\arraystretch}{0.85}
        \begin{tabular}{lccc}
            \toprule
            \textbf{Dataset} & \textbf{Variables} & \textbf{Total Time Steps} & \textbf{Sampling Interval} \\
            \midrule
            ETTh1            & 7                  & 17,420                    & 1 hour                     \\
            ETTh2            & 7                  & 17,420                    & 1 hour                     \\
            ETTm1            & 7                  & 69,680                    & 15 minutes                 \\
            ETTm2            & 7                  & 69,680                    & 15 minutes                 \\
            ECL              & 321                & 26,304                    & 1 hour                     \\
            Exchange         & 8                  & 7,588                     & 1 day                      \\
            Weather          & 21                 & 52,696                    & 10 minutes                 \\
            PEMS03           & 358                & 26,209                    & 5 minutes                  \\
            PEMS07           & 883                & 28,224                    & 5 minutes                  \\
            PEMS08           & 170                & 17,856                    & 5 minutes                  \\
            Illness          & 7                  & 966                       & 1 week                     \\
            \bottomrule
        \end{tabular}}
\end{table}


\subsection{Effectiveness}

We conduct effectiveness experiments on \model~with input sequence $L=96$ and forecast horizons $P \in \{96, 192, $ $336, 720\}$ (PEMS datasets use $P \in \{12, 24, 48, 96\}$, Illness uses $P \in \{24, 36, 48, 60\}$). Table~\ref{tab:effectiveness} presents MSE and MAE comparisons across eleven datasets. \model\ achieves the best overall average MSE or MAE on ETTh1, ETTh2, Weather, and Illness, and ranks second on ETTm1, ETTm2, and ECL. On ECL, \model\ attains the best MSE at horizons 192 and 336 and ties for best MAE at 336, while S-Mamba leads at the shortest and longest horizons to claim the best overall average. On the PEMS traffic benchmarks, \model\ achieves top-two MSE at short prediction lengths (12, 24) across all three subsets and extends this advantage to horizon 48 on PEMS08. On low-variate-count datasets with pronounced periodicity (ETT benchmarks, Illness), \model\ matches or outperforms PatchTST and S-Mamba; the geometric pathway's advantage remains competitive on datasets with richer inter-variable correlation structure, particularly ECL and PEMS. These results confirm that \model\ offers strong, geometry-aware forecasting performance on datasets with both sparse correlations and those with informative, evolving inter-variable correlation dynamics. Low-rank covariance projection preserves these benefits at scale.

The observed performance patterns reflect the geometry-aware dual-path design. Path A models the evolving cross-variable correlation structure as a Riemannian trajectory on the SPD manifold, with the tangent space projection via the log-Euclidean Fr\'echet mean providing a geometrically principled reference whose contribution scales with the richness of inter-variable dynamics, as evidenced on ECL and the PEMS benchmarks. Path B injects this geometric signal directly into the selective state-space parameters ($B$, $C$) of the Mamba VC through the Geometry-Modulated Mamba mechanism, making the SSM's input/output projections responsive to spatial topology changes. The sparse window alignment preserves temporal resolution and avoids information loss during cross-path information transfer.

\begin{table*}[htb!]
    \vspace{-0.1in}
    \caption{Comparison results between \model\ and baselines on 11 datasets in effectiveness experiments. \best{Bold} denotes best MSE/MAE and \second{underline} denotes second-best. The experimental results of baselines on National Illness are derived from experiments. Other baseline results sourced from \cite{wang2025mamba}.}
    \label{tab:effectiveness}
    \renewcommand{\arraystretch}{0.82}
    \centering
    \resizebox{\textwidth}{!}{
        \scriptsize
        \setlength{\tabcolsep}{6pt}
        \setlength{\aboverulesep}{1pt}
        \setlength{\belowrulesep}{1.2pt}
        \vspace{1mm}
        \begin{tabular}{c|c|cc|cc|cc|cc|cc|cc|cc|cc|cc|}
            \toprule
            \multicolumn{2}{c|}{Models}               &
            \multicolumn{2}{c|}{\textbf{\model}}      &
            \multicolumn{2}{c|}{\makecell[c]{\textbf{S-Mamba}                                                                                                                                                                                                                                                                                                                                                                     \\(2025)}} &
            \multicolumn{2}{c|}{\makecell[c]{\textbf{iTransformer}                                                                                                                                                                                                                                                                                                                                                                \\(2024)}} &
            \multicolumn{2}{c|}{\makecell[c]{\textbf{PatchTST}                                                                                                                                                                                                                                                                                                                                                                    \\(2023)}} &
            \multicolumn{2}{c|}{\makecell[c]{\textbf{Crossformer}                                                                                                                                                                                                                                                                                                                                                                 \\(2023)}} &
            \multicolumn{2}{c|}{\makecell[c]{\textbf{TiDE}                                                                                                                                                                                                                                                                                                                                                                        \\(2023)}} &
            \multicolumn{2}{c|}{\makecell[c]{\textbf{TimesNet}                                                                                                                                                                                                                                                                                                                                                                    \\(2023)}} &
            \multicolumn{2}{c|}{\makecell[c]{\textbf{DLinear}                                                                                                                                                                                                                                                                                                                                                                     \\(2023)}} &
            \multicolumn{2}{c|}{\makecell[c]{\textbf{Autoformer}                                                                                                                                                                                                                                                                                                                                                                  \\(2021)}} \\

            \cmidrule(lr){3-4} \cmidrule(lr){5-6} \cmidrule(lr){7-8} \cmidrule(lr){9-10} \cmidrule(lr){11-12} \cmidrule(lr){13-14} \cmidrule(lr){15-16} \cmidrule(lr){17-18} \cmidrule(lr){19-20}
            \multicolumn{2}{c|}{Metric}               & MSE & MAE                        & MSE                        & MAE                        & MSE                       & MAE                      & MSE                      & MAE                      & MSE                      & MAE                      & MSE   & MAE   & MSE   & MAE            & MSE            & MAE            & MSE            & MAE           \\
            \toprule
            \multirow{5}{*}{\rotatebox{90}{ETTm1}}    & 96  & \best{0.329}               & \best{0.367}               & \second{0.333}             & \second{0.368}            & 0.334                    & \second{0.368}           & \best{0.329}             & \best{0.367}             & 0.404                    & 0.426 & 0.364 & 0.387 & 0.338          & 0.375          & 0.345          & 0.372          & 0.505 & 0.475 \\  
                                                      & 192 & \second{0.372}             & 0.392                      & 0.376                      & 0.390                     & 0.377                    & 0.391                    & \best{0.367}             & \best{0.385}             & 0.450                    & 0.451 & 0.398 & 0.404 & 0.374          & \second{0.387} & 0.380          & 0.389          & 0.553 & 0.496 \\  
                                                      & 336 & \second{0.407}             & 0.413                      & 0.408                      & 0.413                     & 0.426                    & 0.420                    & \best{0.399}             & \best{0.410}             & 0.532                    & 0.515 & 0.428 & 0.425 & 0.410          & \second{0.411} & 0.413          & 0.413          & 0.621 & 0.537 \\  
                                                      & 720 & \second{0.471}             & 0.451                      & 0.475                      & \second{0.448}            & 0.491                    & 0.459                    & \best{0.454}             & \best{0.439}             & 0.666                    & 0.589 & 0.487 & 0.461 & 0.478          & 0.450          & 0.474          & 0.453          & 0.671 & 0.561 \\  
            \cmidrule(lr){2-20}
                                                      & Avg & \second{0.395}             & 0.406                      & 0.398                      & \second{0.405}            & 0.407                    & 0.409                    & \best{0.387}             & \best{0.400}             & 0.513                    & 0.495 & 0.419 & 0.419 & 0.400          & 0.406          & 0.403          & 0.407          & 0.588 & 0.517 \\  
            \midrule
            \multirow{5}{*}{\rotatebox{90}{ETTm2}}    & 96  & \second{0.179}             & 0.264                      & \second{0.179}             & \second{0.263}            & 0.180                    & 0.264                    & \best{0.175}             & \best{0.259}             & 0.287                    & 0.366 & 0.207 & 0.305 & 0.187          & 0.267          & 0.193          & 0.292          & 0.255 & 0.339 \\  
                                                      & 192 & \second{0.246}             & \second{0.307}             & 0.250                      & 0.309                     & 0.250                    & 0.309                    & \best{0.241}             & \best{0.302}             & 0.414                    & 0.492 & 0.290 & 0.364 & 0.249          & 0.309          & 0.284          & 0.362          & 0.281 & 0.340 \\  
                                                      & 336 & \second{0.308}             & \second{0.345}             & 0.312                      & 0.349                     & 0.311                    & 0.348                    & \best{0.305}             & \best{0.343}             & 0.597                    & 0.542 & 0.377 & 0.422 & 0.321          & 0.351          & 0.369          & 0.427          & 0.339 & 0.372 \\  
                                                      & 720 & \second{0.407}             & \second{0.403}             & 0.411                      & 0.406                     & 0.412                    & 0.407                    & \best{0.402}             & \best{0.400}             & 1.730                    & 1.042 & 0.558 & 0.524 & 0.408          & \second{0.403} & 0.554          & 0.522          & 0.433 & 0.432 \\  
            \cmidrule(lr){2-20}
                                                      & Avg & \second{0.285}             & \second{0.330}             & 0.288                      & 0.332                     & 0.288                    & 0.332                    & \best{0.281}             & \best{0.326}             & 0.757                    & 0.611 & 0.358 & 0.404 & 0.291          & 0.333          & 0.350          & 0.401          & 0.327 & 0.371 \\  
            \midrule
            \multirow{5}{*}{\rotatebox{90}{ETTh1}}    & 96  & \best{0.384}               & 0.409                      & \second{0.386}             & 0.405                     & \second{0.386}           & 0.405                    & 0.414                    & 0.419                    & 0.423                    & 0.448 & 0.479 & 0.464 & \best{0.384}   & \second{0.402} & \second{0.386} & \best{0.400}   & 0.449 & 0.459 \\  
                                                      & 192 & 0.442                      & 0.440                      & 0.443                      & 0.437                     & 0.441                    & 0.436                    & 0.460                    & 0.445                    & 0.471                    & 0.474 & 0.525 & 0.492 & \best{0.436}   & \best{0.429}   & \second{0.437} & \second{0.432} & 0.500 & 0.482 \\  
                                                      & 336 & \second{0.484}             & 0.461                      & 0.489                      & 0.468                     & 0.487                    & \best{0.458}             & 0.501                    & 0.466                    & 0.570                    & 0.546 & 0.565 & 0.515 & 0.491          & 0.469          & \best{0.481}   & \second{0.459} & 0.521 & 0.496 \\  
                                                      & 720 & 0.504                      & 0.495                      & \second{0.502}             & \second{0.489}            & 0.503                    & 0.491                    & \best{0.500}             & \best{0.488}             & 0.653                    & 0.621 & 0.594 & 0.558 & 0.521          & 0.500          & 0.519          & 0.516          & 0.514 & 0.512 \\  
            \cmidrule(lr){2-20}
                                                      & Avg & \best{0.454}               & 0.451                      & \second{0.455}             & \second{0.450}            & \best{0.454}             & \best{0.448}             & 0.469                    & 0.455                    & 0.529                    & 0.522 & 0.541 & 0.507 & 0.458          & \second{0.450} & 0.456          & 0.452          & 0.496 & 0.487 \\  
            \midrule
            \multirow{5}{*}{\rotatebox{90}{ETTh2}}    & 96  & \best{0.294}               & \best{0.347}               & \second{0.296}             & \second{0.348}            & 0.297                    & 0.349                    & 0.302                    & \second{0.348}           & 0.745                    & 0.584 & 0.400 & 0.440 & 0.340          & 0.374          & 0.333          & 0.387          & 0.346 & 0.388 \\  
                                                      & 192 & \best{0.369}               & \best{0.394}               & \second{0.376}             & \second{0.396}            & 0.380                    & 0.400                    & 0.388                    & 0.400                    & 0.877                    & 0.656 & 0.528 & 0.509 & 0.402          & 0.414          & 0.477          & 0.476          & 0.456 & 0.452 \\  
                                                      & 336 & \best{0.400}               & \best{0.419}               & \second{0.424}             & \second{0.431}            & 0.428                    & 0.432                    & 0.426                    & 0.433                    & 1.043                    & 0.731 & 0.643 & 0.571 & 0.452          & 0.452          & 0.594          & 0.541          & 0.482 & 0.486 \\  
                                                      & 720 & \best{0.406}               & \best{0.433}               & \second{0.426}             & \second{0.444}            & 0.427                    & 0.445                    & 0.431                    & 0.446                    & 1.104                    & 0.763 & 0.874 & 0.679 & 0.462          & 0.468          & 0.831          & 0.657          & 0.515 & 0.511 \\  
            \cmidrule(lr){2-20}
                                                      & Avg & \best{0.367}               & \best{0.398}               & \second{0.381}             & \second{0.405}            & 0.383                    & 0.406                    & 0.387                    & 0.407                    & 0.942                    & 0.683 & 0.611 & 0.550 & 0.414          & 0.427          & 0.559          & 0.515          & 0.450 & 0.459 \\  
            \midrule
            \multirow{5}{*}{\rotatebox{90}{ECL}}      & 96  & \second{0.143}             & \second{0.239}             & \best{0.139}               & \best{0.235}              & 0.148                    & 0.240                    & 0.181                    & 0.270                    & 0.219                    & 0.314 & 0.237 & 0.329 & 0.168          & 0.272          & 0.197          & 0.282          & 0.201 & 0.317 \\  
                                                      & 192 & \best{0.158}               & \second{0.255}             & \second{0.159}             & \second{0.255}            & 0.162                    & \best{0.253}             & 0.188                    & 0.274                    & 0.231                    & 0.322 & 0.236 & 0.330 & 0.184          & 0.289          & 0.196          & 0.285          & 0.222 & 0.334 \\  
                                                      & 336 & \best{0.175}               & \best{0.269}               & \second{0.176}             & \second{0.272}            & 0.178                    & \best{0.269}             & 0.204                    & 0.293                    & 0.246                    & 0.337 & 0.249 & 0.344 & 0.198          & 0.300          & 0.209          & 0.301          & 0.231 & 0.338 \\  
                                                      & 720 & \second{0.209}             & \second{0.302}             & \best{0.204}               & \best{0.298}              & 0.225                    & 0.317                    & 0.246                    & 0.324                    & 0.280                    & 0.363 & 0.284 & 0.373 & 0.220          & 0.320          & 0.245          & 0.333          & 0.254 & 0.361 \\  
            \cmidrule(lr){2-20}
                                                      & Avg & \second{0.171}             & \second{0.266}             & \best{0.169}               & \best{0.265}              & 0.178                    & 0.270                    & 0.205                    & 0.290                    & 0.244                    & 0.334 & 0.252 & 0.344 & 0.193          & 0.295          & 0.212          & 0.300          & 0.227 & 0.338 \\  
            \midrule
            \multirow{5}{*}{\rotatebox{90}{Exchange}} & 96  & \best{0.086}               & \second{0.206}             & \best{0.086}               & 0.207                     & \best{0.086}             & \second{0.206}           & \second{0.088}           & \best{0.205}             & 0.256                    & 0.367 & 0.094 & 0.218 & 0.107          & 0.234          & \second{0.088} & 0.218          & 0.197 & 0.323 \\  
                                                      & 192 & 0.179                      & \second{0.301}             & 0.182                      & 0.304                     & \second{0.177}           & \best{0.299}             & \best{0.176}             & \best{0.299}             & 0.470                    & 0.509 & 0.184 & 0.307 & 0.226          & 0.344          & \best{0.176}   & 0.315          & 0.300 & 0.369 \\  
                                                      & 336 & 0.318                      & \second{0.409}             & 0.332                      & 0.418                     & 0.331                    & 0.417                    & \best{0.301}             & \best{0.397}             & 1.268                    & 0.883 & 0.349 & 0.431 & 0.367          & 0.448          & \second{0.313} & 0.427          & 0.509 & 0.524 \\  
                                                      & 720 & 0.865                      & 0.700                      & 0.867                      & 0.703                     & \second{0.847}           & \best{0.691}             & 0.901                    & 0.714                    & 1.767                    & 1.068 & 0.852 & 0.698 & 0.964          & 0.746          & \best{0.839}   & \second{0.695} & 1.447 & 0.941 \\  
            \cmidrule(lr){2-20}
                                                      & Avg & 0.362                      & \second{0.404}             & 0.367                      & 0.408                     & \second{0.360}           & \best{0.403}             & 0.366                    & \second{0.404}           & 0.940                    & 0.707 & 0.370 & 0.413 & 0.416          & 0.443          & \best{0.354}   & 0.414          & 0.613 & 0.539 \\  
            \midrule
            \multirow{5}{*}{\rotatebox{90}{Weather}}  & 96  & \second{0.166}             & \best{0.209}               & 0.169                      & \second{0.210}            & 0.174                    & 0.214                    & 0.177                    & 0.218                    & \best{0.158}             & 0.230 & 0.202 & 0.261 & 0.172          & 0.220          & 0.196          & 0.255          & 0.266 & 0.336 \\  
                                                      & 192 & \second{0.212}             & \best{0.253}               & 0.214                      & \best{0.253}              & 0.221                    & \second{0.254}           & 0.225                    & 0.259                    & \best{0.206}             & 0.277 & 0.242 & 0.298 & 0.219          & 0.261          & 0.237          & 0.296          & 0.307 & 0.367 \\  
                                                      & 336 & \best{0.272}               & \best{0.296}               & \second{0.274}             & \best{0.296}              & 0.278                    & \best{0.296}             & 0.278                    & \second{0.297}           & \best{0.272}             & 0.335 & 0.287 & 0.335 & 0.280          & 0.306          & 0.283          & 0.335          & 0.359 & 0.395 \\  
                                                      & 720 & \second{0.350}             & \best{0.345}               & 0.353                      & 0.348                     & 0.358                    & \second{0.347}           & 0.354                    & 0.348                    & 0.398                    & 0.418 & 0.351 & 0.386 & 0.365          & 0.359          & \best{0.345}   & 0.381          & 0.419 & 0.428 \\  
            \cmidrule(lr){2-20}
                                                      & Avg & \best{0.250}               & \best{0.276}               & \second{0.253}             & \second{0.277}            & 0.258                    & 0.278                    & 0.259                    & 0.280                    & 0.259                    & 0.315 & 0.270 & 0.320 & 0.259          & 0.286          & 0.265          & 0.317          & 0.338 & 0.382 \\  
            \midrule
            \multirow{5}{*}{\rotatebox{90}{PEMS03}}   & 12  & \second{0.067}             & \second{0.173}             & \best{0.065}               & \best{0.169}              & 0.071                    & 0.174                    & 0.099                    & 0.216                    & 0.090                    & 0.203 & 0.178 & 0.305 & 0.085          & 0.192          & 0.290          & 0.378          & 0.884 & 0.711 \\  
                                                      & 24  & \second{0.092}             & 0.204                      & \best{0.087}               & \best{0.196}              & 0.093                    & \second{0.201}           & 0.142                    & 0.259                    & 0.121                    & 0.240 & 0.257 & 0.371 & 0.118          & 0.223          & 0.320          & 0.398          & 0.834 & 0.692 \\  
                                                      & 48  & 0.142                      & 0.257                      & \second{0.133}             & \second{0.243}            & \best{0.125}             & \best{0.236}             & 0.211                    & 0.319                    & 0.202                    & 0.317 & 0.379 & 0.463 & 0.155          & 0.260          & 0.353          & 0.415          & 0.941 & 0.723 \\  
                                                      & 96  & 0.223                      & 0.327                      & \second{0.201}             & \second{0.305}            & \best{0.164}             & \best{0.275}             & 0.269                    & 0.370                    & 0.262                    & 0.367 & 0.490 & 0.539 & 0.228          & 0.317          & 0.356          & 0.413          & 0.882 & 0.717 \\  
            \cmidrule(lr){2-20}
                                                      & Avg & 0.131                      & 0.240                      & \second{0.121}             & \second{0.228}            & \best{0.113}             & \best{0.222}             & 0.180                    & 0.291                    & 0.169                    & 0.282 & 0.326 & 0.419 & 0.146          & 0.248          & 0.330          & 0.401          & 0.885 & 0.711 \\  
            \midrule
            \multirow{5}{*}{\rotatebox{90}{PEMS07}}   & 12  & \best{0.062}               & \second{0.163}             & \second{0.063}             & \best{0.159}              & 0.067                    & 0.165                    & 0.095                    & 0.207                    & 0.094                    & 0.200 & 0.173 & 0.304 & 0.082          & 0.181          & 0.148          & 0.272          & 0.424 & 0.491 \\  
                                                      & 24  & \second{0.085}             & 0.192                      & \best{0.081}               & \best{0.183}              & 0.088                    & \second{0.190}           & 0.150                    & 0.262                    & 0.139                    & 0.247 & 0.271 & 0.383 & 0.101          & 0.204          & 0.224          & 0.340          & 0.459 & 0.509 \\  
                                                      & 48  & 0.121                      & 0.231                      & \best{0.093}               & \best{0.192}              & \second{0.110}           & \second{0.215}           & 0.253                    & 0.340                    & 0.311                    & 0.369 & 0.446 & 0.495 & 0.134          & 0.238          & 0.355          & 0.437          & 0.646 & 0.610 \\  
                                                      & 96  & 0.183                      & 0.293                      & \best{0.117}               & \best{0.217}              & \second{0.139}           & \second{0.245}           & 0.346                    & 0.404                    & 0.396                    & 0.442 & 0.628 & 0.577 & 0.181          & 0.279          & 0.452          & 0.504          & 0.912 & 0.748 \\  
            \cmidrule(lr){2-20}
                                                      & Avg & 0.113                      & 0.220                      & \best{0.088}               & \best{0.188}              & \second{0.101}           & \second{0.204}           & 0.211                    & 0.303                    & 0.235                    & 0.315 & 0.380 & 0.440 & 0.124          & 0.226          & 0.295          & 0.388          & 0.610 & 0.590 \\  
            \midrule
            \multirow{5}{*}{\rotatebox{90}{PEMS08}}   & 12  & \second{0.077}             & \second{0.180}             & \best{0.076}               & \best{0.178}              & 0.079                    & 0.182                    & 0.168                    & 0.232                    & 0.165                    & 0.214 & 0.227 & 0.343 & 0.112          & 0.212          & 0.154          & 0.276          & 0.436 & 0.485 \\  
                                                      & 24  & \second{0.109}             & \second{0.216}             & \best{0.104}               & \best{0.209}              & 0.115                    & 0.219                    & 0.224                    & 0.281                    & 0.215                    & 0.260 & 0.318 & 0.409 & 0.141          & 0.238          & 0.248          & 0.353          & 0.467 & 0.502 \\  
                                                      & 48  & \second{0.178}             & 0.276                      & \best{0.167}               & \best{0.228}              & 0.186                    & \second{0.235}           & 0.321                    & 0.354                    & 0.315                    & 0.355 & 0.497 & 0.510 & 0.198          & 0.283          & 0.440          & 0.470          & 0.966 & 0.733 \\  
                                                      & 96  & 0.287                      & 0.348                      & \second{0.245}             & \second{0.280}            & \best{0.221}             & \best{0.267}             & 0.408                    & 0.417                    & 0.377                    & 0.397 & 0.721 & 0.592 & 0.320          & 0.351          & 0.674          & 0.565          & 1.385 & 0.915 \\  
            \cmidrule(lr){2-20}
                                                      & Avg & 0.163                      & 0.255                      & \best{0.148}               & \best{0.224}              & \second{0.150}           & \second{0.226}           & 0.280                    & 0.321                    & 0.268                    & 0.306 & 0.441 & 0.464 & 0.193          & 0.271          & 0.379          & 0.416          & 0.814 & 0.659 \\  
            \midrule
            \multirow{5}{*}{\rotatebox{90}{Illness}}  & 24  & \best{1.312}               & \best{0.724}               & 1.707                      & \second{0.792}            & \second{1.530}           & 0.796                    & 2.342                    & 0.982                    & 4.395                    & 1.429 & 2.534 & 1.016 & 2.333          & 0.918          & 3.421          & 1.289          & 4.127 & 1.445 \\  
                                                      & 36  & \best{1.103}               & \best{0.677}               & \second{1.318}             & \second{0.734}            & 1.498                    & 0.769                    & 2.274                    & 0.968                    & 4.096                    & 1.324 & 2.491 & 0.978 & 1.541          & 0.777          & 2.859          & 1.129          & 3.323 & 1.222 \\  
                                                      & 48  & \best{1.180}               & \best{0.698}               & \second{1.432}             & \second{0.777}            & 1.456                    & 0.783                    & 2.125                    & 0.935                    & 4.044                    & 1.308 & 2.331 & 0.958 & 1.502          & 0.787          & 2.633          & 1.089          & 3.258 & 1.224 \\  
                                                      & 60  & \best{1.559}               & \best{0.795}               & 1.930                      & 0.918                     & 1.843                    & 0.887                    & 2.329                    & 0.999                    & 4.296                    & 1.357 & 2.492 & 1.004 & \second{1.788} & \second{0.875} & 2.937          & 1.174          & 3.476 & 1.281 \\  
            \cmidrule(lr){2-20}
                                                      & Avg & \best{1.288}               & \best{0.724}               & 1.597                      & \second{0.805}            & \second{1.582}           & 0.809                    & 2.268                    & 0.971                    & 4.208                    & 1.355 & 2.462 & 0.989 & 1.791          & 0.839          & 2.962          & 1.170          & 3.546 & 1.293 \\  
            \midrule
            \multirow{1}{*}{}                         & --  & \multicolumn{2}{c|}{36/38} & \multicolumn{2}{c|}{30/44} & \multicolumn{2}{c|}{18/21} & \multicolumn{2}{c|}{27/4} & \multicolumn{2}{c|}{3/0} & \multicolumn{2}{c|}{0/0} & \multicolumn{2}{c|}{3/7} & \multicolumn{2}{c|}{6/7} & \multicolumn{2}{c|}{0/0}                                                                                                             \\
            \bottomrule
        \end{tabular}
    }
    \vspace{-0.1in}
\end{table*}


\begin{table*}[!t]
    \vspace{-0.2in}
    \caption{
        Ablation study results for \model\ across ETTh2, ETTm1, Weather, ECL, and PEMS08 datasets. \best{Bold} denotes the best result within each metric and prediction horizon. Prediction lengths:12/24/48/96 for PEMS08, 96/192/336/720 for others.
    }
    \label{tab:ablation}
    \renewcommand{\arraystretch}{0.95}
    \centering
    \resizebox{\textwidth}{!}{
        \setlength{\tabcolsep}{12pt}
        \setlength{\aboverulesep}{1pt}
        \setlength{\belowrulesep}{1pt}
        \footnotesize
        \begin{tabular}{l|c|cc|cc|cc|cc|cc}
            \toprule
            \rule{0pt}{8pt}
            \multirow{2}{*}{Variant}
                                                                          & \multirow{2}{*}{Len}
                                                                          & \multicolumn{2}{c|}{\textbf{ETTh2}}
                                                                          & \multicolumn{2}{c|}{\textbf{ETTm1}}
                                                                          & \multicolumn{2}{c|}{\textbf{Weather}}
                                                                          & \multicolumn{2}{c|}{\textbf{ECL}}
                                                                          & \multicolumn{2}{c}{\textbf{PEMS08}}
            \\
            \cline{3-4}\cline{5-6}\cline{7-8}\cline{9-10}\cline{11-12}
                                                                          &                                       & \rule{0pt}{8pt} MSE & MAE          & MSE          & MAE          & MSE          & MAE          & MSE          & MAE          & MSE          & MAE          \\
            \midrule
            \multirow{5}{*}{\parbox{2.6cm}{Full Model}}                   & 96/12                                 & 0.300               & 0.351        & \best{0.333} & \best{0.368} & \best{0.168} & 0.211        & \best{0.157} & \best{0.251} & \best{0.077} & \best{0.180} \\
                                                                          & 192/24                                & \best{0.382}        & \best{0.401} & 0.379        & 0.395        & 0.217        & 0.257        & \best{0.174} & \best{0.265} & \best{0.108} & \best{0.215} \\
                                                                          & 336/48                                & 0.410               & \best{0.427} & \best{0.411} & \best{0.416} & 0.274        & 0.297        & \best{0.193} & \best{0.286} & \best{0.177} & \best{0.276} \\
                                                                          & 720/96                                & 0.430               & 0.447        & \best{0.478} & 0.455        & 0.350        & 0.347        & \best{0.236} & \best{0.322} & \best{0.294} & \best{0.351} \\
            \cmidrule(r){2-12}
                                                                          & Avg                                   & 0.381               & 0.407        & \best{0.400} & \best{0.409} & \best{0.252} & \best{0.278} & \best{0.190} & \best{0.281} & \best{0.164} & \best{0.256} \\
            \midrule
            \multirow{5}{*}{\parbox{2.6cm}{$\alpha\cdot\tanh$ Injection}} & 96/12                                 & \best{0.298}        & \best{0.349} & 0.336        & 0.370        & 0.169        & 0.213        & 0.185        & 0.274        & 0.078        & 0.181        \\
                                                                          & 192/24                                & 0.383               & 0.402        & 0.381        & 0.397        & 0.219        & 0.256        & 0.193        & 0.283        & 0.110        & 0.216        \\
                                                                          & 336/48                                & \best{0.408}        & 0.428        & 0.412        & 0.417        & 0.280        & 0.299        & 0.227        & 0.318        & 0.191        & 0.286        \\
                                                                          & 720/96                                & 0.427               & 0.446        & \best{0.478} & \best{0.454} & 0.356        & 0.349        & 0.266        & 0.347        & 0.314        & 0.370        \\
            \cmidrule(r){2-12}
                                                                          & Avg                                   & \best{0.379}        & \best{0.406} & 0.402        & 0.410        & 0.256        & 0.279        & 0.218        & 0.306        & 0.173        & 0.263        \\
            \midrule
            \multirow{5}{*}{\parbox{2.6cm}{w/o $B$+$C$ Injection}}        & 96/12                                 & \best{0.298}        & \best{0.349} & 0.337        & 0.370        & 0.169        & 0.212        & 0.181        & 0.270        & 0.079        & 0.182        \\
                                                                          & 192/24                                & 0.383               & 0.402        & 0.383        & 0.399        & 0.219        & 0.256        & 0.193        & 0.283        & 0.113        & 0.218        \\
                                                                          & 336/48                                & \best{0.408}        & \best{0.427} & 0.412        & 0.417        & 0.281        & 0.299        & 0.217        & 0.308        & 0.188        & 0.284        \\
                                                                          & 720/96                                & 0.425               & 0.445        & \best{0.478} & 0.455        & 0.356        & 0.349        & 0.266        & 0.347        & 0.322        & 0.374        \\
            \cmidrule(r){2-12}
                                                                          & Avg                                   & \best{0.379}        & \best{0.406} & 0.402        & 0.410        & 0.256        & 0.279        & 0.214        & 0.302        & 0.176        & 0.265        \\
            \midrule
            \multirow{5}{*}{\parbox{2.6cm}{with $\Delta$ Modulation}}     & 96/12                                 & 0.308               & 0.356        & 0.344        & 0.375        & 0.176        & 0.221        & 0.175        & 0.268        & \best{0.077} & 0.181        \\
                                                                          & 192/24                                & 0.383               & \best{0.401} & 0.401        & 0.409        & 0.217        & 0.257        & 0.192        & 0.284        & 0.115        & 0.221        \\
                                                                          & 336/48                                & 0.412               & \best{0.427} & 0.420        & 0.423        & \best{0.273} & \best{0.296} & 0.215        & 0.308        & 0.194        & 0.293        \\
                                                                          & 720/96                                & \best{0.420}        & \best{0.442} & 0.506        & 0.472        & \best{0.349} & \best{0.346} & 0.263        & 0.346        & 0.304        & 0.354        \\
            \cmidrule(r){2-12}
                                                                          & Avg                                   & 0.381               & 0.407        & 0.418        & 0.420        & 0.254        & 0.280        & 0.211        & 0.302        & 0.173        & 0.262        \\
            \midrule
            \multirow{5}{*}{\parbox{2.6cm}{Linear Interpolation}}         & 96/12                                 & 0.301               & 0.351        & 0.345        & 0.376        & \best{0.168} & \best{0.210} & 0.177        & 0.268        & 0.079        & 0.183        \\
                                                                          & 192/24                                & 0.383               & \best{0.401} & \best{0.378} & \best{0.393} & 0.218        & \best{0.255} & 0.191        & 0.282        & 0.112        & 0.216        \\
                                                                          & 336/48                                & 0.415               & \best{0.427} & 0.413        & 0.419        & 0.277        & 0.298        & 0.215        & 0.307        & 0.197        & 0.288        \\
                                                                          & 720/96                                & 0.427               & 0.445        & 0.480        & 0.456        & 0.353        & 0.348        & 0.262        & 0.345        & 0.323        & 0.373        \\
            \cmidrule(r){2-12}
                                                                          & Avg                                   & 0.382               & \best{0.406} & 0.404        & 0.411        & 0.254        & \best{0.278} & 0.211        & 0.301        & 0.178        & 0.265        \\
            \midrule
            \multirow{5}{*}{\parbox{2.6cm}{Geodesic Smoothness Reg}}      & 96/12                                 & 0.300               & 0.351        & 0.342        & 0.375        & 0.171        & 0.215        & 0.188        & 0.279        & \best{0.077} & 0.181        \\
                                                                          & 192/24                                & \best{0.382}        & \best{0.401} & 0.403        & 0.411        & \best{0.216} & 0.256        & 0.193        & 0.283        & 0.110        & \best{0.215} \\
                                                                          & 336/48                                & 0.410               & \best{0.427} & 0.428        & 0.429        & 0.274        & 0.297        & 0.216        & 0.307        & 0.188        & 0.285        \\
                                                                          & 720/96                                & 0.430               & 0.447        & 0.484        & 0.460        & 0.350        & 0.347        & 0.265        & 0.347        & 0.295        & \best{0.351} \\
            \cmidrule(r){2-12}
                                                                          & Avg                                   & 0.381               & 0.407        & 0.414        & 0.419        & 0.253        & 0.279        & 0.216        & 0.304        & 0.167        & 0.258        \\
            \bottomrule
        \end{tabular}
    }
    \vspace{0in}
\end{table*}


\subsection{Ablation Study}
\label{sec:ablation}

Table~\ref{tab:ablation} presents comprehensive ablation results on ETTh2, ETTm1, Weather, ECL, and PEMS08 datasets with variable counts ranging from 7 to 321. For PEMS08, prediction lengths are $\{12, 24, 48, 96\}$, and they are $\{96,192,$ $336,720\}$ for the other datasets. We evaluate six architectural variants to assess the contribution of geometric injection design and alignment strategies:

\vpara{Geometric Injection Variants:} (i)~\textbf{Full Model}: direct additive injection of geometric signals into $B$ and $C$ (default). (ii)~\textbf{$\alpha \cdot \tanh$ Injection}: applies a learnable scale factor $\alpha$ with $\tanh$ bounding to the geometric modulation, providing stability through bounded $(-1, 1)$ modulation. (iii)~\textbf{$\Delta$ Modulation}: additionally injects geometry into the time-step size $\boldsymbol{\Delta}$ through an MLP alongside $B$ and $C$ injection, enabling the geometric trajectory to influence the SSM's forgetting rate. (iv)~\textbf{w/o $B$+$C$ Injection}: removes all geometric modulation to isolate the contribution of the geometry pathway.

The Full Model achieves the best average MSE and MAE on four of five datasets, confirming that direct, unbounded additive $B$+$C$ injection is the most effective configuration. Removing $B$+$C$ injection reveals a pronounced gradient of geometric dependency across dataset dimensionalities: on ECL ($N = 321$), the average MSE degrades by \textbf{12.6\%} (0.190 $\to$ 0.214), with the largest gap at horizon 96 (0.157 $\to$ 0.181, +15.3\%); on PEMS08 ($N = 170$), the degradation is \textbf{7.3\%} (0.164 $\to$ 0.176). In contrast, on low-variate-count datasets such as ETTh2 and ETTm1, removing geometric injection causes negligible performance change (e.g., ETTh2: 0.381 $\to$ 0.379), indicating that the geometric trajectory provides limited additional information when the inter-variable correlation space is small. This dimensional threshold effect validates the core design hypothesis: the geometric injection is most impactful in high-dimensional datasets where rich inter-variable correlation dynamics exist to be modeled.

The $\alpha \cdot \tanh$ variant yields marginal gains on ETTh2 (average MSE 0.379 vs.\ 0.381), with improvements concentrated at shorter horizons. However, on ECL its performance degrades substantially (0.190 $\to$ 0.218, +14.7\%), demonstrating that bounding the geometric modulation via $\tanh$ restricts the injection magnitude when strong geometric signals are needed to modulate SSM parameters in high-dimensional regimes. The $\Delta$ Modulation variant exhibits a complementary trade-off: while it underperforms the Full Model on ETTm1 and ECL, it achieves the best results at the longest horizon on ETTh2 (H720: 0.430 $\to$ 0.420) and ties the Full Model on Weather at long horizons (H336: 0.274 $\to$ 0.273; H720: 0.350 $\to$ 0.349). This suggests that modulating the forgetting rate $\boldsymbol{\Delta}$ provides complementary temporal control that becomes beneficial specifically at extended prediction horizons, though at the cost of increased optimization difficulty in high-dimensional settings.

\vpara{Alignment and Regularization:} (v)~\textbf{Linear Interpolation} replaces the sparse window alignment with linear interpolation to map geometric signals in temporal windows to the variate grid. (vi)~\textbf{Geodesic Smoothness Regularization} adds an auxiliary loss penalizing large deviations between consecutive log-covariance matrices to encourage smoother geometric trajectories on the SPD manifold.

The default sparse window alignment outperforms linear interpolation in most cases, particularly evident in high-dimensional datasets and long-term prediction: ECL improves by \textbf{11.1\%} (0.211 $\to$ 0.190), PEMS08 by \textbf{8.5\%} (0.178 $\to$ 0.164), while on low-variate-count datasets, where some results prove inferior to the replacement. This confirms that sparse placement, which preserves the native temporal resolution of each covariance window at its exact position, avoids the information smearing introduced by interpolation-based alignment across variable tokens. The Geodesic Smoothness Regularization degrades performance on most datasets (ECL: 0.190 $\to$ 0.216; ETTm1: 0.400 $\to$ 0.414), indicating that the log-Euclidean Fr\'echet mean and tangent space projection inherently enforce sufficient geometric smoothness without explicit auxiliary regularization. The learned covariance trajectory is already constrained by the manifold geometry to evolve coherently, making additional smoothness penalties redundant.

\subsection{Efficiency}
\label{sec:efficiency}

\begin{figure*}[!t]
    \centering
    \includegraphics[width=\linewidth]{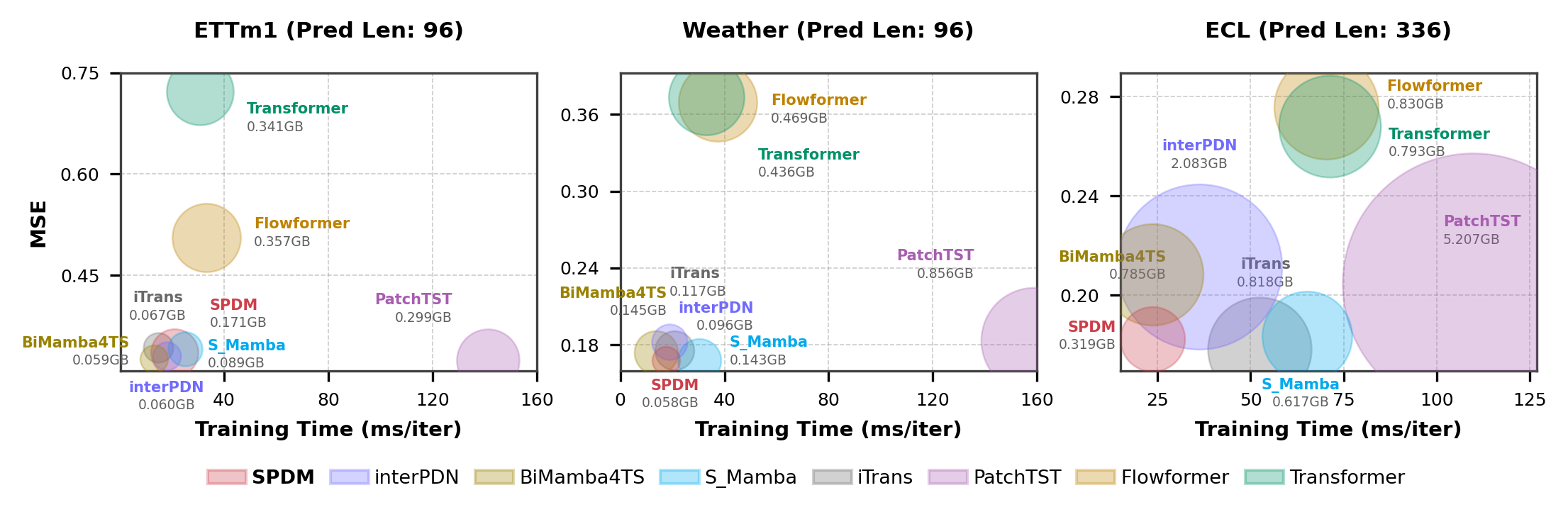}
    \caption{Comparison of efficiency experimental results between \model\ and baselines.}
    \label{fig:efficiency}
\end{figure*}

Figure~\ref{fig:efficiency} presents a comprehensive efficiency comparison between \model\ and representative baselines on ETTm1, Weather, and ECL, spanning forecasting accuracy (MSE), training time (milliseconds per iteration), and maximum GPU memory allocated. All experiments use either unified hyperparameter configurations or those suggested in the official repositories. The main model employs the optimal configuration obtained through hyperparameter search.

\model\ consistently achieves the lowest or near-lowest MSE across all three studies while yielding substantial improvements in both training speed and memory efficiency. In terms of forecasting accuracy, \model\ attains competitive MSE values of approximately 0.335 on ETTm1, 0.168 on Weather, and 0.182 on ECL. On training speed, \model\ completes one iteration in approximately 23 ms on ECL, compared to roughly 36 ms for interPDN and 110 ms for PatchTST, showing its great scalability and efficiency in handling multivariate forecasting. The peak memory of \model\ remains compact: approximately 175 MB on ETTm1 and 326 MB on ECL, demonstrating that incorporating geometric constraints does not introduce significant overhead and maintains \model\'s efficiency when surpassing state-of-the-art models.

This efficiency advantage derives from two complementary design properties. First, Mamba VC encoder is built upon the Mamba SSM, which inherits linear time complexity with respect to sequence length, in contrast to the cost of Transformer and patching baselines. Second, the SPD manifold trajectory path projects high-dimensional cross-variable correlations onto a compact tangent space via the logarithmic map, producing a low-dimensional sequence that the Geometry Mamba processes with minimal overhead. For large-scale datasets, the low-rank covariance projection bounds the SPD operation cost to $\mathcal{O}(R^3)$, ensuring computational tractability at scale.

\subsection{Robustness}

\begin{figure*}[!t]
    \centering
    \includegraphics[width=\textwidth]{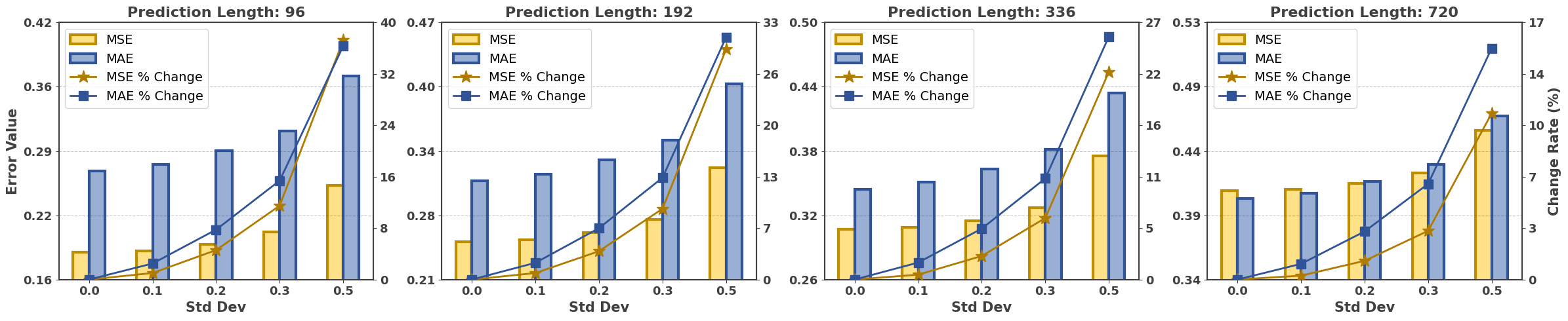}
    \caption{Robustness experimental results on ETTm2 under increasing Gaussian noise.}
    \label{fig:robustness}
\end{figure*}

Figure~\ref{fig:robustness} presents the robustness analysis of \model~on ETTm2, where Gaussian white noise with increasing standard deviation $\sigma \in \{0.0, 0.1, 0.2, 0.3, 0.5\}$ is injected into the input sequences across forecast horizons $P \in \{96, 192, 336,$ $ 720\}$. Both MSE and MAE and their relative percentage changes are reported at each noise level. As $\sigma$ rises, the error metrics exhibit a controlled and gradual upward trend consistent with expectation under increasing input perturbation. Crucially, the degradation rate is tightly constrained: at $\sigma = 0.3$, the MSE increase remains negligible across all horizons; even under strong perturbation at $\sigma = 0.5$, the MSE growth at the longest prediction horizon is limited to approximately 11\%. In stark contrast to models operating directly in the raw input space, which typically suffer performance collapse at comparable noise levels,\model\ maintains functional forecasting accuracy throughout the entire perturbation range.

This noise resilience stems from two mechanisms. First, the per-window covariance estimation aggregates signals across $W$ time steps, attenuating zero-mean random components. Second, the log-Euclidean Fr\'echet mean and tangent space projection act as an implicit geometric regularizer, suppressing high-frequency perturbations while preserving genuine correlation structure changes. The geometric modulation signals injected into the SSM parameters ($B$, $C$) thus reflect genuine topological changes in the cross-variable correlation structure rather than stochastic artifacts.

\subsection{Lookback Length Study}

\begin{figure*}[!t]
    \centering
    \includegraphics[width=\linewidth]{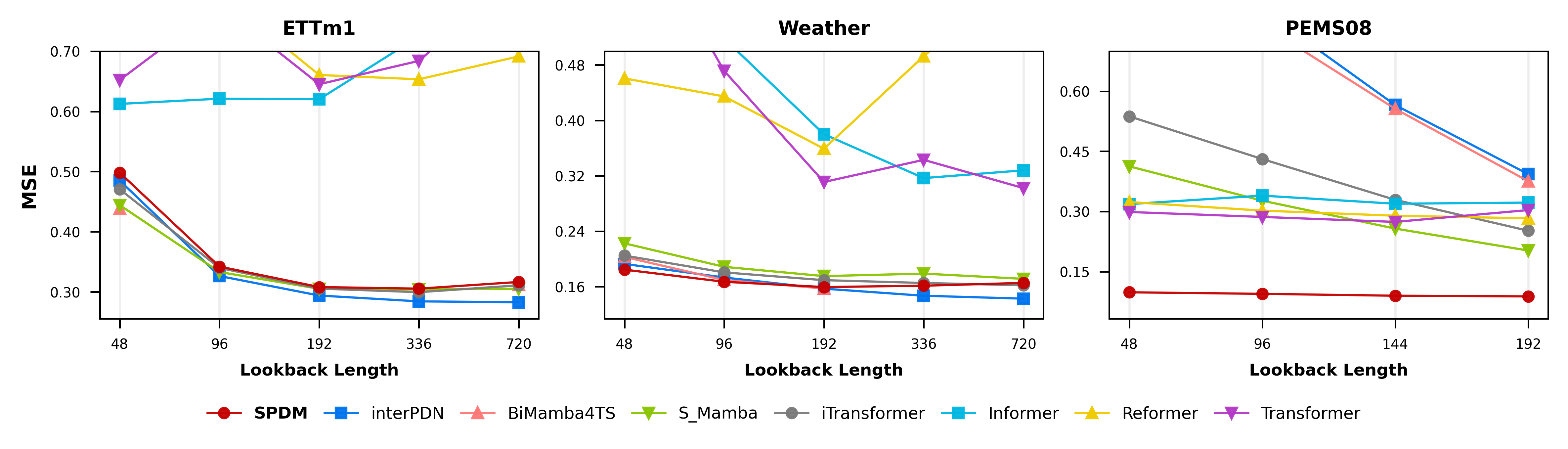}
    \caption{Prediction error of \model\ and baselines with increasing lookback length.}
    \label{fig:lookback}
\end{figure*}

Figure~\ref{fig:lookback} illustrates the effect of progressively increasing the lookback window length on the forecasting accuracy of \model\ and baseline models across three datasets with diverse physical dynamics. The experiments use lookback window ranges of $L \in \{48, 96, 192, 336, 720\}$ for ETTm1 and Weather, and $L \in \{48, 96, 144, 192\}$ for PEMS08 reflecting its shorter temporal granularity. Across all configurations, \model\ exhibits a clear and smooth monotonic decreasing trend in MSE as the lookback window expands, consistently delivering the best or near-best forecasting accuracy over the entire range of window sizes. Notably, on PEMS08, \model\ maintains strong predictive accuracy and its MSE continues to decline as the sequence length increases, without any sign of performance rebound even at the maximal lookback length of 192. In contrast, baseline models such as Transformer and Informer exhibit performance saturation or noticeable degradation: their MSE curves plateau at intermediate windows and in several cases rebound upward at $L = 336$ or $L = 720$, indicating a fundamental inability to effectively exploit longer historical contexts. S-Mamba, while capable of benefiting from extended lookback windows, yields an error curve that remains consistently above that of \model\ across the full spectrum of Weather and ECL.

The robust long-range modeling capability of \model\ is grounded in two interconnected design properties. First, the mature inverted embedding and Mamba VC encoder's structured state transition preserves spatial dependencies with linear-time complexity, enabling \model\ to extract effective patterns from extended historical observations without being overwhelmed by redundant or noisy inputs. Second, the SPD manifold trajectory path provides implicit geometric regularization by mapping the dynamic covariance structure onto $\mathcal{P}_N$ and modulating the hidden state transition through informative constraints, the model effectively suppresses high-frequency stochastic perturbations. The continuous geometric evolution of covariance patterns on the manifold acts as a spatiotemporal smoother, attenuating noise components that lack coherent geometric structure while faithfully preserving the underlying correlation dynamics.

\subsection{Hyperparameter Sensitivity Study}
\label{sec:sensitivity}

\begin{figure}
    \centering
    \includegraphics[width=\linewidth]{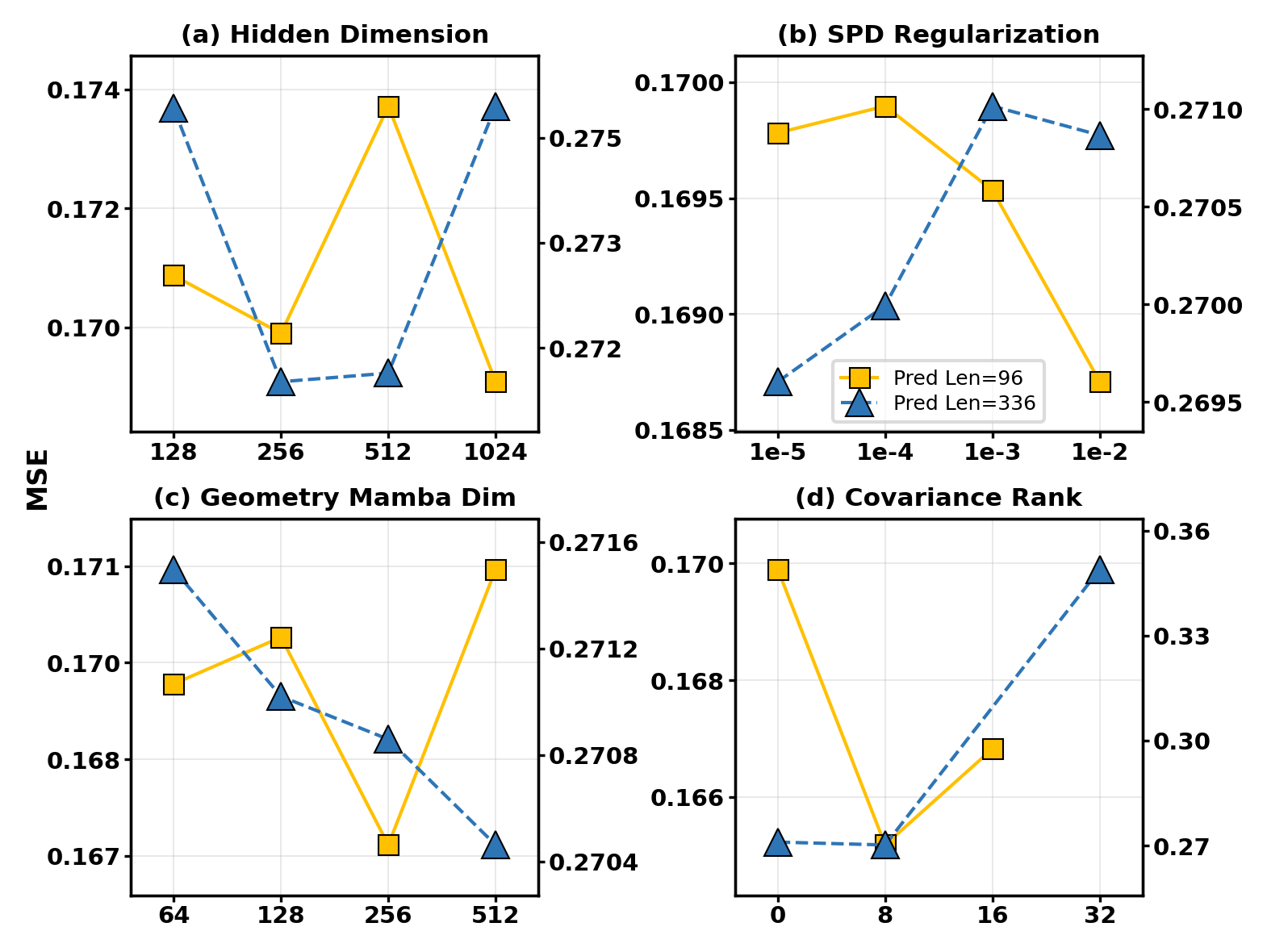}
    \caption{Results of hyperparameter sensitivity study.}
    \label{fig:sensitivity}
\end{figure}

Figure~\ref{fig:sensitivity} examines the sensitivity of \model~to four core hyperparameters through sub-figures spanning prediction length $L \in \{96, 336\}$ on Weather. Those are (a)~Hidden Dimension $d_{\text{model}}$, (b)~SPD Regularization constant $\epsilon$, (c)~Geometry Mamba Dimension $d_{\text{geo}}$, and (d)~Covariance Rank $R$ of the low-rank projection.

Subfigure (a) reveals that moderate hidden dimensions $d_{\text{model}} = 256$ achieve almost the best forecasting accuracy, with excessively large dimensions yielding no additional benefit. Sub-figure~(d) demonstrates that low-rank covariance projection with $R \in \{8, 16\}$ not only substantially reduces the eigendecomposition cost from $\mathcal{O}(N^3)$ to $\mathcal{O}(R^3)$, but also achieves accuracy comparable to or exceeding that of the full-rank counterpart with $N=21$, indicating that the dominant modes of correlation dynamics are well captured within a compact low-dimensional subspace.\model\ also shows remarkably flat sensitivity to $\epsilon$ and $d_{\text{geo}}$. At the prediction length $P = 96$, the MSE fluctuates within a negligible band of 0.001-0.005 across the entire evaluated range of these parameters, demonstrating that the model's accuracy does not rely on meticulous hyperparameter tuning of these components. Notably, for high-dimensional datasets, an excessively small $\epsilon$ may provide insufficient regularization for covariance matrices, rendering them nearly singular and thereby destabilizing the eigen-decomposition and subsequent matrix logarithm on the SPD manifold, which can ultimately cause training failures.

Collectively, these results affirm that \model's superior forecasting performance stems from its foundational geometric inductive bias, the architectural principle of modeling cross-variable correlations as a continuous Riemannian trajectory on the SPD manifold, rather than from exhaustive hyperparameter search or parameter scaling. The insensitivity to $\epsilon$ and $d_{\text{geo}}$ demonstrates that manifold constraints and geometric state-space dynamic modeling provide a geometrically robust framework that tolerates wide variations in its auxiliary parameterization.

\subsection{Case Study}

\begin{figure}
    \centering
    \includegraphics[width=\linewidth]{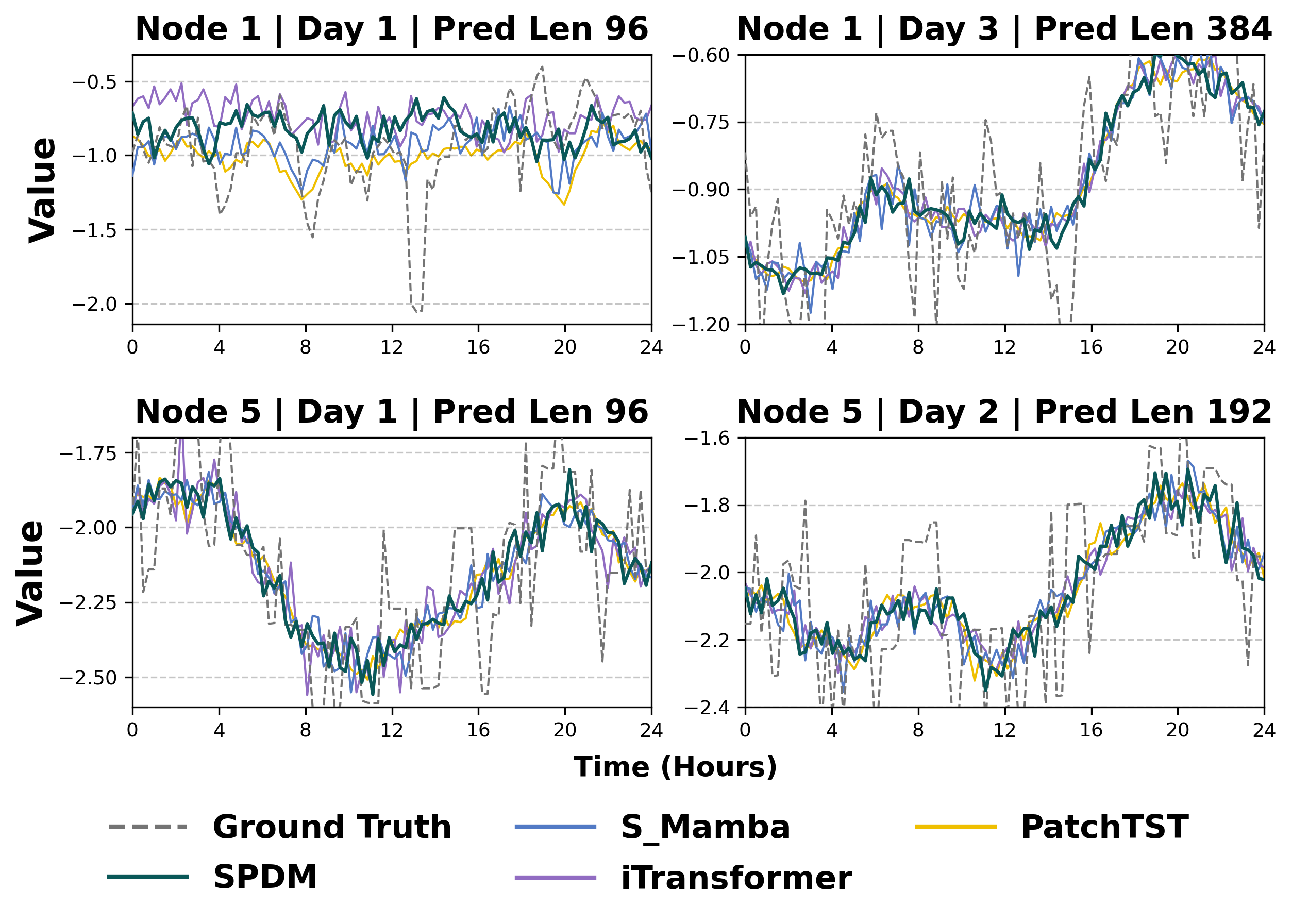}
    \caption{Ground truth and prediction results of \model\ and baselines on ETTm2. Node~1 represents high useful load (HUFL); Node~5 represents low useless load (LULL).}
    \label{fig:case-study-square}
\end{figure}

Figure~\ref{fig:case-study-square} presents the scaled prediction results of \model, S-Mamba, iTransformer, and PatchTST against the ground truth on the ETTm2 dataset, which reflects electricity transformer temperature and load with 15-minute sampling. The study covering two representative nodes across four forecasting scenarios: Node~1 with prediction lengths 96 (Day~1) and 384 (Day~3) represents high useful load, and Node~5 with prediction lengths 96 (Day~1) and 192 (Day~2) represents low useless load. These sub-figures reveal two complementary capabilities that distinguish \model\ from competing architectures: noise resilience against high-frequency localized perturbations, and responsiveness to long-range structural regime shifts.

\vpara{Noise Resilience.} The ground-truth signal of Node~1 on Day~1 exhibits sharp, localized downward spikes typical of electricity transformer data under sudden load fluctuations. Baselines such as iTransformer and PatchTST are misled by these stochastic excursions, producing erroneous fluctuations that overfit to the instantaneous noise. In contrast, \model\ successfully suppresses these spurious transients and closely adheres to the underlying signal. This is the direct consequence of Path~A's geometric processing: per-window covariance estimation attenuates zero-mean random components, while the log-Euclidean manifold projection at the Fr\'echet mean functions as an implicit low-pass filter in the geometric domain.

\vpara{Structural Regime-Shift Tracking.} Node~1 on Day~3 exhibits a pronounced baseline elevation spanning approximately the 14th through 18th hour. Baseline models like S-Mamba display a visible lag in responding to this shift, whereas \model\ tracks the transition seamlessly and settles into the post-shift regime without delay. Node~5 further reveals \model's precision at complex nonlinear valley dynamics: at inflection points where the signal transitions from descent to ascent, competing models exhibit phase shifts or oversmoothing to varying degrees, while \model\ achieves competitive spatio-temporal alignment. These capabilities derive from the geometric modulation mechanism. When the cross-variable covariance trajectory on $\mathcal{P}_N$ experiences a geometric deflection, the Geometry Mamba encodes this evolution, and Mamba VC injects such dynamics to hidden state transition procedures, enabling the SSM's memory dynamics to adapt to the changing correlation topology.

\section{Conclusion}
\label{sec:conclusion}
This paper introduces the principle of state-space dynamic modeling with manifold constraints for multivariate time series forecasting. By treating the evolving cross-variable correlation structure as a continuous Riemannian trajectory on the SPD manifold and directly injecting geometric signals into the selective parameters of Mamba VC, \model\ demonstrates that manifold constraints serve as principled geometric regularization, simultaneously improving forecasting accuracy and noise resilience while preserving the linear-time complexity of the Mamba parallel scan.

Experiments on eleven real-world benchmarks establish state-of-the-art forecasting performance, with \model\ achieving the best or near-best average MSE on datasets with varied physical characteristics. Systematic ablation reveals that direct additive injection of geometric signals into the SSM's $B$ and $C$ parameters is the dominant performance factor, whose contribution scales with dataset dimensionality; sparse window alignment and the log-Euclidean Fr\'echet mean reference point prove essential. Hyperparameter sensitivity further confirms the model's robustness.

By seamlessly integrating manifold-constrained dynamics with the linear-complexity selective scan, \model\ establishes a paradigm for bridging Riemannian geometry and efficient state-space modeling. Promising extensions include handling irregularly sampled signals, exploiting higher-order product manifolds and Grassmannians for richer correlation representations, and developing self-supervised pre-training strategies that leverage geometric trajectory representations for cross-domain transfer learning.


\bibliographystyle{cas-model2-names}
\bibliography{ref}

\end{document}